\documentclass[11pt]{article}

\usepackage[numbers]{natbib}

\usepackage{graphicx}
\usepackage{amsmath,amssymb,amsthm,amsfonts}
\usepackage{mathtools}
\usepackage{booktabs}
\usepackage{nicefrac}
\usepackage{microtype}
\usepackage{enumitem}
\usepackage[table]{xcolor}
\usepackage{tabularx}
\usepackage{wrapfig}
\usepackage[normalem]{ulem}
\usepackage{bbm}
\usepackage{dsfont}
\usepackage{grffile}
\usepackage{url}
\usepackage{epstopdf}
\usepackage{algorithm}
\usepackage{algpseudocode}
\usepackage[margin=1in]{geometry}
\usepackage{hyperref}
\usepackage{cleveref}

\hypersetup{colorlinks=true,linkcolor=blue,citecolor=teal,filecolor=teal,urlcolor=cyan}

\newtheorem{theorem}{Theorem}[section]
\newtheorem{lemma}[theorem]{Lemma}

\newtheorem{proposition}[theorem]{Proposition}

\newtheorem{remark}[theorem]{Remark}

\newcommand{\best}[1]{\makebox[0pt][l]{\textbf{#1}}\phantom{#1}}
\newcommand{\second}[1]{\uline{#1}}

\DeclareMathOperator{\ELBO}{ELBO}

\newcommand{\N}{\mathbb{N}}  % natural numbers
\newcommand{\R}{\mathbb{R}}  % real numbers
\newcommand{\abs}[1]{\left\lvert#1\right\rvert}  % absolute value
\newcommand{\diff}{\mathop{}\!\mathrm{d}}  % differential
\newcommand{\deriv}[3][]{\frac{\mathrm{d}^{#1}#2}{\mathrm{d} #3^{#1}}}  % derivative
\newcommand{\iid}{\stackrel{\mathrm{iid}}{\sim}}  % iid
\newcommand{\norm}[1]{\left\lVert#1\right\rVert}  % norm
\newcommand{\pderiv}[3][]{\frac{\partial^{#1}#2}{\partial #3^{#1}}}  % partial derivative
\DeclareMathOperator*{\argmin}{arg\,min}

\DeclareMathOperator{\E}{\mathbb{E}}  % expectation
\DeclareMathOperator{\Gaussian}{\mathcal{N}}  % Gaussian
\newcommand{\calD}{\mathcal{D}}

\newcommand{\calL}{\mathcal{L}}

\newcommand{\calN}{\mathcal{N}}

\newcommand{\calX}{\mathcal{X}}
\newcommand{\calY}{\mathcal{Y}}

\title{Diffusion-DFL: Decision-focused Diffusion Models for Stochastic Optimization}
\author{
Zihao Zhao\thanks{\texttt{zzhao628@gatech.edu}. Georgia Institute of Technology.}
\and
Christopher Yeh\thanks{\texttt{cyeh@caltech.edu}. California Institute of Technology
}
\and
Lingkai Kong\thanks{\texttt{lingkaikong@g.harvard.edu}. Harvard University.}
\and
Kai Wang\thanks{\texttt{kwang692@gatech.edu}. Georgia Institute of Technology.}
}
\date{October 6, 2025}

\begin{document}

\begin{titlepage}
\maketitle
\begin{abstract}

Decision-focused learning (DFL) integrates predictive modeling and optimization by training predictors to optimize the downstream decision target rather than merely minimizing prediction error. To date, existing DFL methods typically rely on deterministic point predictions, which are often insufficient to capture the intrinsic stochasticity of real-world environments. To address this challenge, we propose the first diffusion-based DFL approach, which trains a diffusion model to represent the distribution of uncertain parameters and optimizes the decision by solving a stochastic optimization with samples drawn from the diffusion model. Our contributions are twofold. First, we formulate diffusion DFL using the reparameterization trick, enabling end-to-end training through diffusion. While effective, it is memory and compute-intensive due to the need to differentiate through the diffusion sampling process. Second, we propose a lightweight score function estimator that uses only several forward diffusion passes and avoids backpropagation through the sampling. This follows from our results that backpropagating through stochastic optimization can be approximated by a weighted score function formulation. We empirically show that our diffusion DFL approach consistently outperforms strong baselines in decision quality. The source code for all experiments is available \href{https://github.com/GT-KOALA/Diffusion_DFL}{here}.

\end{abstract}
\thispagestyle{empty}
\end{titlepage}

% {\hypersetup{linkcolor=black}\tableofcontents}
% \newpage

\section{Introduction}
Many real-life decision-making tasks require selecting actions that minimize a cost function involving unknown, context-dependent parameters. These parameters must often be predicted from observed features. For example, in supply chain management, future product demand must be estimated before deciding how much inventory to order~\citep{tang_identifying_2011}. A common approach is the predict-then-optimize pipeline, where a predictive model is first trained using a loss function such as mean squared error (MSE), and the resulting predictions are then passed to an optimization solver to guide decisions. While simple and widely adopted, this two-stage method can be misaligned with the true objective: minimizing decision cost. In particular, lower prediction error does not always lead to higher-quality decisions~\citep{bertsimas_predictive_2020,elmachtoub_smart_2022}.

Decision-focused learning (DFL) addresses this misalignment by integrating the prediction and optimization stages into a single end-to-end framework~\citep{donti_task-based_2017, wilder_melding_2019, mandi_decision-focused_2024}. Unlike the two-stage approach, DFL trains the prediction model specifically to improve decision outcomes, often resulting in solutions with lower regret. However, most existing DFL methods rely on point (deterministic) predictions as inputs to the optimization layer, despite the fact that in many real-world scenarios, the underlying parameters are inherently uncertain and may follow complex distributions. Ignoring this uncertainty can lead to overconfident models and degraded decision quality~\citep{kochenderfer_decision_2015}.

In this work, we introduce a novel DFL approach that leverages diffusion probabilistic models to capture the environment uncertainty in an end-to-end fashion. Here, we use a conditional diffusion model~\citep{tashiro_csdi_2021} to represent the distribution of uncertain parameters given contextual features. The advantage of integrating a diffusion model into DFL is that, unlike simple distribution predictions (e.g., Gaussian), diffusion models can capture multi-modal or complex distributions. However, the sequential sampling procedure of diffusion models introduces a challenge when training a diffusion model end-to-end for stochastic optimization. To address this, we develop two algorithms: reparameterization and score function. First, the reparameterization trick is a common approach that expresses a random sample as a deterministic function of the model parameters and some noise, and we can backpropagate through sampled prediction to solve the DFL problem. 

However, this approach can be very costly in memory and computation because it requires differentiating (and therefore tracking gradients) through the diffusion sampling process. To address this, we introduce a lightweight score function estimator that avoids differentiating through the sampling process. Specifically, we use a score function surrogate to approximate the gradient of the diffusion predictor and plug it into the KKT (Karush-Kuhn-Tucker) implicit-differentiation approach to obtain the total derivative of the decision objective.
% for each sample, our method computes the gradient with a single forward pass of the diffusion model's denoising loss},
% yielding a substantially more memory-efficient procedure. 
In addition, we further mitigate the high variance that arises from using only score functions for a few steps by employing a tailored importance sampling strategy.

We evaluate our proposed methods in various applications, including (synthetic) product allocation, energy scheduling, and stock portfolio optimization. Experimental results show that our diffusion DFL methods consistently outperform all baselines, with more improvements on larger problem sizes. Moreover, the score function estimator achieves decision quality comparable to that of the reparameterization method, while significantly reducing GPU memory usage from 60.75 GB to 0.13 GB. The contributions of this paper are the following:
%An alternative is to use score function gradient estimators, which avoid differentiating through the sampling process. The score function method provides a gradient of the expected decision loss by weighting each sample's outcome with the gradient of the log-probability of the sample (the score). This technique does not require any internal derivatives of the predictor, making it broadly applicable even when the decision model is not differentiable. We also prove that in this case, the training loss for the diffusion model is equivalent to a weighted denoising loss, which can be easily calculated and implemented. \lk{This paragraph is almost discussing the general score function method. However, score function is very well-known. We should emphasize when using score function for diffusion model training, what is the novelty?} 
% \kai{Not coherent with the previous sentence. Please add some transition to your summary below.}
\begin{itemize}[left=1em,nosep]
    \item We introduce the first DFL method that uses diffusion models to capture the downstream uncertainty and employs the reparameterization trick for end-to-end gradient estimation.
    \item We propose a lightweight score function estimator that avoids backpropagating the reversing process in the reparameterization method, significantly reducing memory and computation cost.
    \item We evaluate our methods in three real-world optimization tasks and observe consistent improvements over strong baselines.
\end{itemize}
\section{Related Works}
\paragraph{Decision-focused learning}
DFL is an emerging paradigm that 
%integrates predictive modeling with downstream optimization, 
trains models end-to-end to directly optimize decision quality rather than minimizing prediction error~\citep{donti_task-based_2017, wilder_melding_2019, mandi_decision-focused_2024}. 
% This approach contrasts with the conventional \emph{predict-then-optimize} pipeline, where a predictive model is trained in isolation; instead, DFL incorporates the structure and objectives of the decision problem into the learning process~\citep{elmachtoub_smart_2022}. 
% A number of studies have demonstrated the superior decision performance achieved by DFL methods~\citep{wang_automatically_2020}, and we refer readers to a recent survey for a comprehensive background~\citep{mandi_decision-focused_2024}
Despite the success in aligning learning objectives with decision-making, a limitation of most existing DFL methods is that they typically rely on \textbf{deterministic point predictions} of uncertain parameters~\citep{wilder_melding_2019, shah_decision-focused_2022}. By ignoring distributional uncertainty, deterministic point predictions cannot represent the full outcomes and may lead to lower decision quality~\citep{wang_gen-dfl_2025}. Empirically, classic DFL was observed to struggle in high-dimensional and risk-sensitive real-world settings with significant uncertainty~\citep{mandi_decision-focused_2022}.

Therefore, the gap in uncertainty modeling motivates the need for more comprehensive DFL with \textit{stochastic predictions}, where several works have started integrating uncertainty awareness into the DFL pipeline~\citep{silvestri_score_2023,wang_gen-dfl_2025, shariatmadar_generalized_2025, jeon_locally_2025}. 
For instance, \cite{wang_gen-dfl_2025} proposes a generative DFL approach (Gen-DFL) based on normalizing flow models as the predictor. 
%For instance, \cite{silvestri_score_2023} introduced to learn simple probabilistic models (e.g., assuming a Gaussian distribution over outcomes) to capture the uncertainty instead of just fixed points.
However, normalizing flows require a bijective network architecture, which restricts the expressiveness of the stochastic predictor.

In this paper, we propose using diffusion models~\citep{ho_denoising_2020} as a more expressive predictor. By leveraging diffusion models in the DFL paradigm, our approach extends DFL by predicting an accurate full distribution of the unknown parameters, which addresses the overconfidence of deterministic optimization and better aligns with downstream decision-making needs.

\paragraph{Diffusion model in optimization}
Diffusion probabilistic models have achieved great success in modeling high-dimensional data distributions in recent years~\citep{sohl-dickstein_deep_2015, song_generative_2019, dhariwal_diffusion_2021}. Originally popularized for image generation and related structured outputs, its ability to capture multi-modal and high-variety distributions has made it attractive beyond vision tasks, such as combinatorial optimization\citep{sun_difusco_2023, sanokowski_diffusion_2025}, black-box optimization\citep{krishnamoorthy_diffusion_2023, kong_diffusion_2025}.
To our best knowledge, however, no prior work has integrated diffusion models into a predict-then-optimize learning pipeline for decision tasks. This paper is the first to harness diffusion models in an end-to-end DFL framework. By using a conditional diffusion model, we can learn a rich distribution over the uncertain inputs and then propagate this uncertainty through to the downstream decision via gradient-based training (score function and reparameterization). This approach combines the strengths of expressive generative modeling and DFL to improve decision quality under uncertainty.
\section{Problem Statement}
\subsection{Decision-focused learning}
% We consider a general predict-then-optimize setting \citep{donti_task-based_2017,elmachtoub_smart_2022} where a decision $z \in \R^d$ is made under uncertainty in some \textit{problem parameter} $y \in \calY$. The cost of using decision $z$ when the true problem parameter has value $y^*$ is evaluated by a decision loss function $f: \calY \times \R^d \to \R$. Assume that a given feature vector $x \in \calX$ is associated with each ground-truth problem parameter $y^*$, with joint distribution $\calD$ over $\calX \times \calY$. 

% In DFL, we first optimize to obtain decisions by sampling from a simulated predictor, and then evaluate the decisions in the real environment and update the predictor. The overall goal is to learn a decision function $z^*_\theta: \calX \to \R^d$, with \textit{model parameter} $\theta$, that minimizes the expected decision loss under ground-truth problem parameters:
% \begin{equation}
% \label{eq:upper-level}
%     \min_\theta F(\theta) := \E_{(x, y^*) \sim \calD} [f(y^*, z_\theta(x))].
% \end{equation}

We consider a general predict-then-optimize setting \citep{donti_task-based_2017,elmachtoub_smart_2022}, where the goal is to make decisions under uncertainty about a key problem parameter. Given a feature vector $x \in \calX$ and a prediction of an unknown parameter $y^* \in \calY$, the decision-maker selects $z \in \R^d$ to minimize a decision loss function $f: \calY \times \R^d \to \R$, which measures the cost of applying decision $z$ when the true parameter is $y^*$. We assume a joint distribution $\calD$ over $(x, y^*)$ pairs.

DFL integrates prediction and optimization into a unified framework. The goal is to learn a decision function $z_\theta^*: \calX \to \R^d$, parameterized by $\theta$, that minimizes the expected decision loss,
\begin{equation}
\label{eq:upper-level}
    \min_\theta F(\theta) := \E_{(x, y^*) \sim \calD} [f(y^*, z^*_\theta(x))].
\end{equation}
The decision $z^*_\theta(x)$ is typically obtained by solving an optimization problem involving a prediction of the uncertainty parameter. Most DFL methods~\citep{mandi_decision-focused_2024} use a deterministic point prediction $y_\theta(x)$ of the uncertain parameter $y^*$:
\begin{equation}
    z_\theta^*(x) = \argmin_z f(y_\theta(x), z), \quad\text{s.t.} \ \ Gz \leq h,\ A z = b,
\end{equation}
where $G \in \R^{n\times d}, h \in \R^{n}, A \in \R^{p \times d}, b \in \R^p$ are constraint problem coefficients\footnote{We consider affine constraints in our main paper for simplicity. The extension from affine constraints to general convex constraints $h(x, z) \leq 0$ follows a similar derivation as in the linear case.}. 

In contrast, we consider a probabilistic model $P_\theta(\cdot \mid x)$ for the uncertain parameter $y^*$ and let $z_\theta^*(x)$ be the solution to a stochastic optimization problem:
\begin{equation}\label{eq:z_star_sto}
    z_\theta^*(x) = \argmin_z \E_{y \sim P_\theta(\cdot|x)} [f(y,z)], \quad\text{s.t.} \ \ Gz \leq h,\ A z = b.
\end{equation}
We aim to learn the model parameter $\theta$ such that $z_\theta^*$ minimizes the expected decision loss $F(\theta)$. By the chain rule, the derivative of $F$ is
\begin{equation*}
    \deriv{F(\theta)}{\theta}
    = \E_{(x,y^*) \sim \calD}\left[ \pderiv{f(y^*, z_\theta^*(x))}{z} \deriv{z_\theta^*(x)}{\theta} \right].
\end{equation*}
However, computing this gradient (specifically, the $\deriv{z_\theta^*}{\theta}$ term) is challenging because $z_\theta^*$ is implicitly defined by a nested optimization problem. 
A common solution is to differentiate the KKT system that implicitly defines $z^*_\theta$ w.r.t. $\theta$~\citep{amos_optnet_2017}.
Another crucial point is the selection of the stochastic predictor in DFL, which in the paper we choose to use diffusion models to represent $P_\theta$.

\subsection{Diffusion probabilistic model}
\label{sec:pre_diffusion}
To generate complex multi-modal and high-dimensional distributions,
diffusion probabilistic models~\citep{ho_denoising_2020} are a promising way. It couples a fixed noising chain with a learned reverse denoising chain.  Let $y_0 \in \R^d$ denote a sample from the real data distribution $q(y_0)$ and $\{\beta_t \in (0, 1)\}_{t=1}^T$ denote the noise schedule. Define $\alpha_t = 1 -\beta_t$ and $\bar\alpha_t = \prod_{i=1}^t \alpha_i$. The \emph{forward process} $q$ adds Gaussian noise at each step $t$ to $y_1$ through $y_T$:
\begin{align}
    q(y_t \mid y_{t-1}) = \mathcal{N}(y_t; \sqrt{1 - \beta_t}y_{t-1}, \beta_t I),
        \quad t=1, \dots, T,
\end{align}
which guarantees that $q(y_T \mid y_0)$ becomes nearly standard normal distribution as $T \to \infty$ with common schedules ($\bar \alpha_T \to 0$). Note that $y_t$ can be equivalently sampled without iterating through intermediate time steps: $y_t = \sqrt{\bar \alpha_t} y_0 - \sqrt{1 - \bar \alpha_t}\epsilon$, where $\epsilon \sim N(0, I)$ is a Gaussian noise.

In the \emph{reverse process} $p$, the diffusion model predicts the unknown added noise by
\begin{align}\label{eq:diffusion_reverse}
    p_\theta(y_{t-1} \mid y_t) = \calN(y_{t-1};\ \mu_\theta(y_t, t),\, \sigma_t^2I),
\end{align}
whose mean $\mu_\theta(\cdot, t)$ is parameterized by a neural network predictor and variance is either fixed ($\sigma_t^2 = \beta_t$) or learned. 
%A common parameterization predicts the added noise $\epsilon_\theta(y_t, t)$ and sets $\mu_\theta (y_t, t) = \frac{1}{\sqrt{\alpha_t}} (y_t - \frac{\beta_t}{\sqrt{1 - \bar \alpha_t}}\epsilon_\theta(y_t, t))$. In diffusion training, we draw $t \sim \text{Unif}(1, \dots, T)$ and $\epsilon \sim \calN(0, I)$, forward construct $y_t = \sqrt{\bar \alpha_t}y_0 + \sqrt{1 - \bar \alpha_t}\epsilon$, and reverse minimize a (weighted) MSE loss $\| \epsilon - \epsilon_\theta(y_t, t) \|^2$. For more details, we refer the readers to~\citet{ho_denoising_2020}.
The combination of $p$ and $q$ is equivalent to a hierarchical variational auto-encoder~\citep{vahdat_nvae_2020}, and thus can be optimized by using the evidence lower bound (ELBO) as the loss function~\citep{hoffman_elbo_2016}. 
% We postpone the discussion for this to \Cref{sec:ll_to_elbo}.

\textbf{Conditional Diffusion Model.} Throughout this paper, $x$ denotes contextual features, and every transition probability is conditioned on $x$~\citep{tashiro_csdi_2021}. 
% This induces the conditional generative model:
% \begin{align}
%     P_\theta(y \mid x) \equiv p_\theta(y_0 \mid x) = \int p(y_T)\prod_{t=1}^{T} p_\theta(y_{t-1} \mid y_t, x)\, dy_{1:T},
% \end{align}
% with $p(y_T) = \calN(y_T;\ 0, I)$. 
We use $P_\theta(\cdot|x)$ for the diffusion model's conditional distribution for generated data and $p_\theta(y_{t-1} \mid y_t, x)$ for its Markov transitions.

\section{Stochastic Optimization and Reparameterization estimator}\label{sec:reparm}

\begin{figure}[htbp]
    \centering
    \includegraphics[width=1\linewidth]{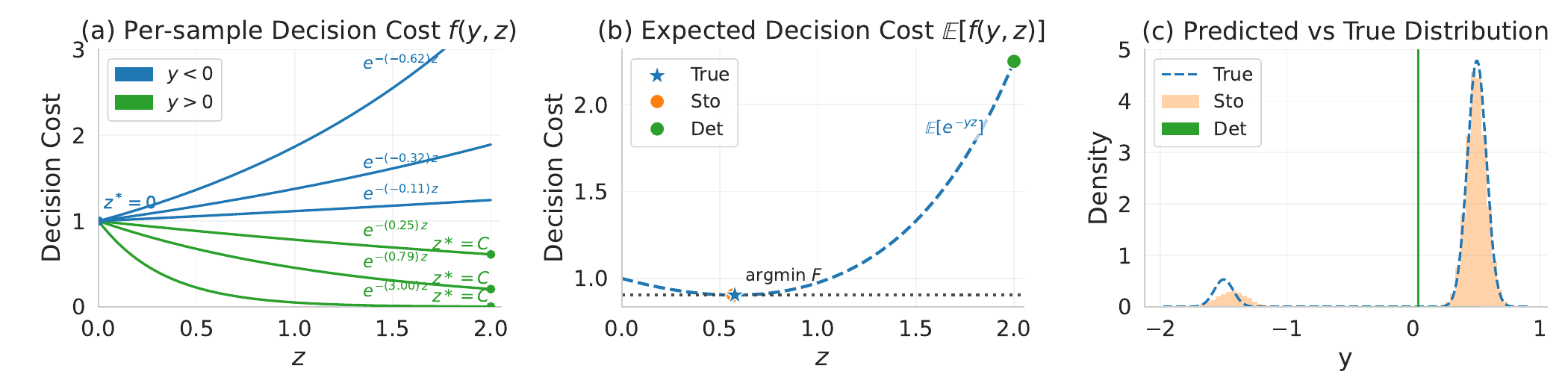}
    \vspace{-0.7cm}
    \caption{A comparison of deterministic vs. stochastic optimization with cost function $\exp(-yz)$, as described in \Cref{sec:exp_toy}. \textbf{(a)} Each curve represents a cost function given a sample $y$. For any fixed $y$, the deterministic optimization decision lies at one of the boundaries ($z^*=0$ or $z^*=C$). \textbf{(b)} When averaging the cost function over many samples of $y$, the stochastic optimization decision lies in the interior of the feasible region instead of on the boundary. Thus, any deterministic optimization decision is suboptimal. \textbf{(c)} A probabilistic (diffusion) model captures a distribution over $Y$ that closely resembles the true bimodal distribution.}
    \label{fig:motivation_plot}
    \vspace{-0.1cm}
\end{figure}

%In a deterministic approach, the model provides a single point prediction for uncertain parameters. 
Real-world decision problems often face significant uncertainty in their parameters. 
%which makes \emph{stochastic optimization} crucial~\citep{bertsimas_predictive_2020}. 
Optimizing with a stochastic predictor (e.g., diffusion model) yields better results than deterministic optimization,
by explicitly modeling the uncertainty and optimizing the expected cost. 
\Cref{fig:motivation_plot} illustrates a simple example: any deterministic solution ends up at an extreme decision with a higher expected cost, while the stochastic solution averages costs across likely outcomes and selects an interior decision with a lower expected cost.

%In other words, the decision is chosen to perform well on average instead of a single point. 

\paragraph{Solving stochastic DFL.} Formally, in the stochastic case, the optimality condition for the decision problem must consider an expectation. The stationarity condition for decision problem Eq.~\ref{eq:z_star_sto} becomes:
\begin{align}
    \nabla_z \calL (\theta, z^*, \lambda^*, \nu^*; x) = \E_{y \sim P_\theta(\cdot \mid x)} [\nabla_z f(y,z)] + G^\top \lambda^{*} + A^\top \nu^{*} = 0,
\end{align}
where $\calL$ denotes the Lagrangian. Note that the dependency on $\theta$ in the stationarity condition is in the distribution. Therefore, we need to handle this dependency carefully while differentiating the KKT system with respect to $\theta$:
\begin{align}\label{eq:deriv_lagrange}
    \underbrace{\frac{\mathrm{d}}{\mathrm{d} \theta} (\nabla_z \calL (\theta, z^*, \lambda^*, \nu^*; x))}_{\text{distributional gradient}} = \frac{\mathrm{d}}{\mathrm{d} \theta} (\E_{y \sim P_\theta(\cdot|x)} [\nabla_z f( y,z)] + G^\top \lambda^{*} + A^\top \nu^{*}) = 0.
\end{align}
To resolve the dependence of both the predictive distribution $P_\theta(y|x)$ and the decision $z^*$ on $\theta$, we first adopt the \textit{reparameterization trick} \citep{Kingma2014} for the diffusion model. From Section~\ref{sec:pre_diffusion}, recall that the diffusion sampling process introduces Gaussian noise at each step. Thus, we can reparameterize the reverse process by fixing all the random draws (Gaussian noises). Formally, a sample $y \sim P_\theta(y \mid x)$ can be expressed as a transformation $y = R(\epsilon, \theta \mid x)$ of a base Gaussian noise sample $\epsilon \sim P(\epsilon)$, where $R$ is differentiable in $\theta$. This makes the diffusion sampling a deterministic function of $\theta$. Then we have
\begin{align}\label{eq:prop_reparam}
     \nabla_\theta{\E_{y \sim P_\theta(\cdot|x)} [f(y,z)]} = \E_{\epsilon \sim P(\epsilon)} [(\nabla_\theta R(\epsilon, \theta|x))^\top \nabla_y f(y, z)].
\end{align}
% \red{This implies that we can compute the distributional gradient by first sampling $\epsilon$ and then computing $y$ by running the diffusion sampling process, and we can obtain $\nabla_\theta R(\epsilon, \theta|x)$ by backpropagating through the generative process.}

Next, we incorporate this into the optimization. Following Eq.~\ref{eq:deriv_lagrange}, we can formalize a KKT system that contains derivatives of $z^*_\theta, \lambda^*, \nu^*$ and $ \nabla_\theta{\E_{y \sim P_\theta(\cdot|x)} [f(y,z)]}$. Plugging the reparameterized gradient estimator into the KKT system, we can solve for $\deriv{z_\theta^*}{\theta}$ and then obtain the total derivative of the final objective $F$ by multiplying $\frac{dF}{dz_\theta^*}$~\citep{donti_task-based_2017} (proof can be found in~\Cref{apd:proof_rp}):
\begin{align}\label{eq:total_grad_reparam}
    \frac{d F}{d \theta} = - 
    \begin{bmatrix}
    \frac{d F}{d z_\theta^*} \\
    0 \\
    0
    \end{bmatrix}^\top 
    \begin{bmatrix}
    H & G^\top & A^\top \\
    D(\lambda^*) G &  D(Gz_\theta^*-h) & 0\\
    A & 0 & 0
    \end{bmatrix}^{-1} \begin{bmatrix}
    \E_{\epsilon \sim P(\epsilon)} [(\nabla_\theta R(\epsilon, \theta|x))^\top \nabla^2_{zy} f(y, z_\theta^*)]  \\
    0 \\
    0
    \end{bmatrix},
\end{align}
where $H = \E_{y \sim P_\theta(\cdot|x)} [\nabla_{zz}^2 f(y, z_\theta^*)]$ is the Hessian of the Lagrangian with respect to $z$, and $D(v)$ denotes a diagonal matrix with $v$ on its diagonal.
In practice, one can sample $\epsilon$ from a certain distribution (e.g., Gaussian) multiple times to estimate the expectation and then obtain the gradient. This gives us reparameterization-based diffusion DFL using Eq.~\ref{eq:total_grad_reparam} to run stochastic DFL optimization.

\section{Score function estimator}\label{sec:score_fn}
A major obstacle to implementing the total gradient (Eq.~\ref{eq:total_grad_reparam}) is the need to backpropagate through the diffusion sampling process. In most cases, the diffusion model's generative process is complex and multi-step (e.g., 1000 steps), which makes backpropagating through all those steps memory-intensive and prone to instability. To address this, we propose a \textbf{score function}\footnote{In this paper, score function refers to the statistical score $\nabla_\theta \log P_\theta(y|x)$ (gradient of log-likelihood w.r.t. model parameters), as opposed to Stein's score $\nabla_y p(y_{t}|y_{t-1}, x)$ often used in diffusion literature.} gradient estimator for the diffusion model, which circumvents explicit backpropagation through all sampling steps. The key idea is to rewrite the Jacobian $\nabla_\theta y$ in terms of the score $\nabla_\theta \log P_\theta(y\mid x)$, and then approximate the score with the diffusion model’s ELBO training loss.

\subsection{Transform the Jacobian into Score Function}\label{sec:ll_to_elbo}
We begin by rewriting the gradient of expectation as an expectation of a score function using the \emph{log-trick}~\citep{mohamed_monte_2020}. Formally, if $y \sim P_\theta(\cdot|x)$ and $f(y)$ is any function not dependent on $\theta$, then by the log-trick we have
\begin{align}\label{eq:grad_ll}
    \nabla_\theta \E_{y \sim P_\theta(\cdot \mid x)} [f(y, z)]
    = \E_{y \sim P_\theta(\cdot\mid x)} [f(y, z) \cdot \nabla_\theta \log P_\theta(y \mid x)].
\end{align}
Intuitively, instead of differentiating the output $y$ through each diffusion step, we only need to compute the gradient for the final log-likelihood, which avoids the need to differentiate through the diffusion sampling process and yields an efficient estimator for the gradient.

Then, one remaining difficulty is that directly computing the exact $\nabla_\theta \log P_\theta(y|x)$ is complicated in practice because $P_\theta(y|x)$ is defined as the marginal probability of $y$ after integrating out the latent diffusion trajectory. 
To obtain a computationally efficient estimator, we use the diffusion model's training objective as a surrogate for the log-likelihood. Specifically, diffusion models are typically trained by maximizing an ELBO that lower-bounds the log-likelihood:
\begin{align}
    \log P_\theta(y_0) &= \log \int p_\theta(y_0 | y_1)\, p_\theta(y_1 | y_2)\, \cdots p_\theta(y_{T-1} | y_T)\, p_\theta(y_T)\, d y_{1:T} \nonumber\\
    &= \log \E_{y_{t} \sim q(y_{t} | y_{t-1}) \forall t \in [T]} \left[\prod_{t=1}^T \frac{p_\theta(y_{t-1}|y_t)}{q(y_t | y_{t-1})} p_\theta(y_T) \right] \nonumber \\
    &\geq \E_{y_{t} \sim q(y_{t} | y_{t-1}) \forall t \in [T]} \left[ \sum_{t=1}^T \log \frac{q(y_t | y_{t-1})}{p_\theta(y_{t-1} | y_t)} + \log p_\theta(y_T) \right] := \ELBO(y_0; \theta), \nonumber
\end{align}
where the inequality is due to Jensen's. To approximate $\nabla_\theta \log P_\theta(y_0 | x)$ conditioned on $x$, we use the gradient of the conditional ELBO loss as a surrogate:
\begin{align}\label{eq:elbo_grad_approx}
    \nabla_\theta \log P_\theta(y_0|x) \approx \nabla_\theta \ELBO(y_0|x; \theta).
\end{align}
\begin{wrapfigure}{r}{0.42\textwidth}
  \vspace{-0.5cm}
  \centering
  \includegraphics[width=\linewidth]{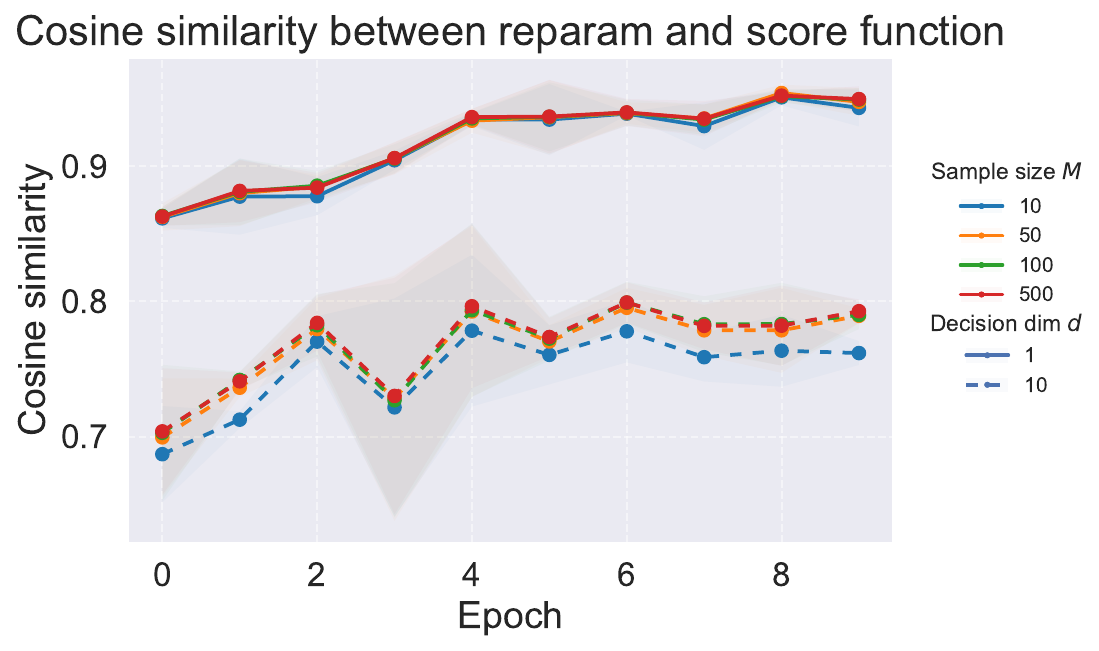}
  \vspace{-0.9cm}
  \caption{Cosine similarity between the reparameterization and score function gradient across different dimensions.}
  % \vspace{-7pt}
  \label{fig:grad_cos}
\end{wrapfigure}
In practice, we first sample a final output $y$ from the diffusion model given contextual features $x$. We then sample a subset of $k$ timesteps $\{t_1, t_2, \dots, t_k\}$ ($k \ll T$) and run forward noising process $q$ to generate the trajectory $\{y_{t_1}, y_{t_2}, \dots, y_{t_k}\}$. As in DDPM~\citep{ho_denoising_2020}, we adopt the simplified form of $\ELBO \approx \E_{t\sim[T], y_0, \epsilon_t}[|\|\epsilon_t - \epsilon_\theta(y_t, t) \|^2]$. We evaluate the $\ELBO$ on the sampled trajectories and compute its gradient w.r.t. $\theta$ as an estimation to the true score.

% Although the ELBO is only a lower bound for log-likelihood, 
Empirical evidence suggests that the ELBO gradient closely tracks the true score, as shown in~\Cref{fig:grad_cos}, making Eq.~\ref{eq:elbo_grad_approx} a reliable proxy in practice. 
% Moreover, our score estimator is fully black-box: it requires no modifications to the diffusion sampler (easy implementation) and scales to long diffusion steps.

\subsection{Overall Gradient for Score Function} \label{sec:discusion_sf}
By plugging the ELBO gradient approximation from Eq.~\ref{eq:elbo_grad_approx} into Eq.~\ref{eq:grad_ll}, we can express the KKT conditions without using reparameterization and thus obtain the score function-based derivative:
% Using the result in Eq~\ref{eq:grad_ll} and the ELBO gradient approximation in Eq.~\ref{eq:elbo_grad_approx}, we can follow the same derivation of the reparameterization method using the KKT system and obtain the total derivative as
\begin{align}\label{eq:full_grad_sf}
    \frac{d F}{d \theta} &\approx - 
    \begin{bmatrix}
    \frac{d F}{d z_\theta^*} \\
    0 \\
    0
    \end{bmatrix}^\top 
    \begin{bmatrix}
    H & G^\top & A^\top \\
    D(\lambda^*) G &  D(Gz_\theta^* - h) & 0\\
    A & 0 & 0
    \end{bmatrix}^{-1} \begin{bmatrix}
    \E_{y \sim P_\theta(\cdot|x)} [\nabla_z f(y, z_\theta^*) (\nabla_\theta{\ELBO(y|x;\theta)}^\top]  \\
    0 \\
    0
    \end{bmatrix}.
\end{align}

\paragraph{Practical algorithm -- weighted ELBO gradient.} 
%The key difference between the reparameterization and score function is the computational efficiency. The reparameterization method requires full backpropagation through the diffusion reverse-sampling process, incurring high GPU memory costs. 
To compute the score surrogate in practice, we found it convenient to treat the total gradient as an importance-weighted form:
\begin{align}\label{eq:weighted_elbo}
    \frac{d F}{d \theta}
    \approx \frac{d}{d \theta} \E_{y \sim P_\theta(\cdot|x)}[ \underbrace{\operatorname{detach}[w_\theta(y)]}_{\text{importance weight, no grad in $\theta$}} \cdot \ \underbrace{{\ELBO (y|x, \theta)}}_{\text{1-step forward}}],
\end{align}
where $w_\theta(y)$ is the importance weight simplified from Eq.~\ref{eq:full_grad_sf} (see~\Cref{apd:prf_IS_elbo} for complete form).
This yields a \emph{weighted-ELBO} gradient estimator: we treat $w_\theta(y)$ as a stop-gradient weight and only differentiate the $\ELBO$ w.r.t. $\theta$, greatly reducing computations. We implement the entire gradient computation as a user-friendly PyTorch autograd module: the forward pass returns the optimal decision $z^*$ (and $\lambda^*, \nu^*$), and the backward pass computes the gradient $\frac{d F}{d \theta}$ as derived above.

\paragraph{Variance-reduction strategy.}While the score-function estimator is effective, a naive implementation of the weighted ELBO loss in Eq.~\ref{eq:weighted_elbo} can suffer from high variance, leading to unstable training. In practice, we found that carefully designing the sampling strategy for the ELBO loss is crucial to obtaining low-variance and stable gradients. To reduce the variance, we utilize the method from Improved DDPM~\citep{nichol_improved_2021} for choosing diffusion steps. Specifically, instead of uniform sampling, we use \emph{importance sampling} over timesteps with probability $p_t$ and weights $1/p_t$:
\begin{align} 
    \nabla_\theta \text{ELBO}^{\text{IS}} = \E_{t \sim p_t} \left[\frac{\nabla_\theta (\text{ELBO}_t)}{p_t}\right], \ \text{where} \ p_t \propto \sqrt{\E [\|\nabla_\theta (\text{ELBO}_{t})\|^2]} \ \text{and} \sum_t p_t=1.
\end{align}
This method remains unbiased, but the variance is minimized.
In essence, this approach gives less weight to the early timesteps that have large gradients and more weight to later timesteps.

% \textbf{Reusing samples by importance sampling:}
% \begin{align}
%     &\E_{y^\prime \sim P_{\theta^\prime}(\cdot|x)} [\nabla_z f(z^*, y^\prime) ELBO(y^\prime, \theta^\prime)] \\
%     &= \E_{y \sim P_{\theta}(\cdot|x)} [\frac{P_{\theta^\prime}(y^\prime|x)}{P_{\theta}(y|x)} \nabla_z f(z^*, y) ELBO(y, \theta^\prime)] \\
%     &= \E_{y \sim P_{\theta}(\cdot|x)} [\exp(\log{P_{\theta^\prime}(y^\prime|x)} - \log{P_{\theta}(y|x)}) \nabla_z f(z^*, y) ELBO(y, \theta^\prime)]
% \end{align}
\section{Experiments}
We evaluate the performance of our diffusion-based DFL approaches on a variety of tasks, comparing against several baseline methods. Specifically, we consider:
\begin{itemize}[nosep,left=0.5em]
    \item \textbf{Two-stage predict-then-optimize baselines}: a deterministic MLP, a Gaussian probabilistic model, and a diffusion model trained to minimize prediction error~\citep{elmachtoub_smart_2022}.
    \item \textbf{Deterministic DFL}: a deterministic MLP model with end-to-end DFL training~\citep{donti_task-based_2017}.
    \item \textbf{Gaussian DFL} (both reparameterization and score function): a Gaussian probabilistic model with end-to-end stochastic DFL training~\citep{silvestri_score_2023}; see details in Appendix~\ref{apd:gs}.
    \item \textbf{Diffusion DFL (ours)}: our diffusion model predictor, trained with either reparameterization or score-function gradient estimators.
\end{itemize} 
% To quantify decision quality, we use the \emph{regret metric}, defined as the expected difference in cost between the decision obtained using the model's prediction and the cost of the optimal decision using the true labels. Formally, if ${z}_\theta(x)^*$ is the decision obtained from model predictions and $z_{\text{true}}^*(x)$ is the optimal decision for true outcome $y$, then the regret is
% \begin{align}
%     \text{Regret}(\theta) = \E_{(x,y) \sim D} [f(z_\theta^*(x), y) - f(z_{\text{true}}^*(x), y)],
% \end{align}
% where lower regret indicates better performance.

\subsection{Synthetic Example}
\label{sec:exp_toy}

In this example, we consider a factory that decides how much to manufacture for each of $d \in \N$ products. The parameter $Y \in \R^d$ represents the \textit{profit margin} for each product, i.e., $Y_i$ is the profit per unit of product $i$; due to uncertainty in market conditions, $Y$ is uncertain. The factory's decision $z \in [0,C]^d$ represents how much of each product to manufacture, where $C$ is the maximum capacity for each product. For simplicity, we do not consider any contextual features $x$ in this example. That means DFL learns a distribution that generates $y$ that can minimize the decision objective.

Suppose that the factory has a risk-averse cost function $f(y,z) = \exp(-y^\top z)$\footnote{Here, we have ignored the degenerate case $y=0$. To deal with the degenerate case, one could add a zero-centered bump function $c(y)$ to the objective $f(y,z)$.}, which indicates that the factory wants to put a larger weight on the product with higher profit $Y_i$. Under uncertainty, the decision-maker seeks to minimize the \textbf{expected cost} by solving a stochastic optimization problem:
\begin{align}
    z_\mathrm{sto}^* &\in \argmin_{z \in [0,C]^d} \E_{y \sim P_\theta(\cdot|x)} [\exp(-y^\top z)].
\end{align}
In this stochastic case, the optimal investment $z_\mathrm{sto}^*$ typically lies in the interior of the feasible region, which balances the potential high reward of investing against the risk of losses.

\textbf{Experimental setup.}
We simulate the uncertain parameter $Y$ drawn from a mixture of Gaussians,
\begin{align}
    Y_i \ \iid \
    p \cdot \Gaussian(a, \sigma^2) + (1-p) \cdot \Gaussian(-b, \sigma^2).
\end{align}
Specifically, we set $p=0.8, a = 1, b=3, \sigma=0.15, C= 2$. %We train each model (deterministic, Gaussian, diffusion) on this distribution and evaluate the expected cost achieved by the decision $z^*$. 
We train each model (deterministic, Gaussian, diffusion) on this distribution in a decision-focused manner (for DFL methods) or on pure regression (for two-stage), and evaluate the expected cost achieved by the resulting decision $z_\theta^*(x)$. We present the results of one product ($d=1$) in~\Cref{fig:motivation_plot} and 10 products ($d=10$) in~\Cref{tab:exp_all_methods}.

\subsection{Power Schedule}
In this experiment, we evaluate our method on a real-world energy scheduling problem from \citet{donti_task-based_2017}. This task involves a 24-hour generation-scheduling problem in which the operator chooses $z \in \R^{24}$ (hourly generation). Given a realization $y$ of demand, the decision loss penalizes shortage and excess with asymmetric linear costs ($\gamma_s$ and $\gamma_e$) plus a quadratic tracking term; the decision must also satisfy a ramping bound $c_r$. Let $[v]_+ := \max(v, 0)$. We have the decision loss as the quadratic problem:
\begin{align}\label{eq:quad_prob}
    &\min_z \  \E_{y \sim P_\theta(\cdot|x)}[f(y, z)] = \sum_{i=1}^{24} \E_{y \sim P_\theta(\cdot|x)} [\gamma_s[y_i - z_i]_+ + \gamma_e[z_i - y_i]_+ + \frac{1}{2} (z_i - y_i)^2], \notag \\
    &\ \mathrm{s.t.} \ |z_i - z_{i - 1}| \leq c_r \ \text{for all} \ i \in \{1, 2, \dots, 24\}.
\end{align}
\textbf{Experimental setup.} We use more than 8 years of historical data from a regional power grid~\citep{pjm_interconnection_data_2025}. Feature $x$ includes the previous day's hourly load, temperature, next-day temperature forecasts, non-linear transforms (lags and rolling statistics), calendar indicators, and yearly sinusoidal features. Given $x$, the prediction model $P_\theta(\cdot|x)$ outputs a distribution over $y \in \mathbb{R}^{24}$. We report the test decision cost in~\Cref{tab:exp_all_methods} and a held-out horizon in~\Cref{fig:power_grid_24_hour}.

\subsection{Stock Market Portfolio Optimization}
In this experiment, we apply our diffusion DFL approach to a financial portfolio optimization problem under uncertain stock returns. Here, the random vector $y \in \R^n$ represents the returns for the assets $n$ on the next day, and the decision $z \in \R^n$ represents the portfolio weights allocated to those assets. We consider a mean-variance trade-off decision loss: maximize expected return while keeping the risk (variance) low. This can be written as minimizing a loss that is a negative expected return plus a quadratic penalty on variance:
\begin{align}
    \min_z \ \E_{y \sim P_\theta(\cdot|x)}[f(y, z)] = \E_{y \sim P_\theta(\cdot|x)} \left[\frac{\alpha}{2} z^\top y y^\top z - y^\top z\right], \quad \mathrm{s.t.}\quad z^\top \mathbf{1} = 1, \ 0 \leq z_i \leq 1,
\end{align}
where $\alpha > 0$ is a risk parameter and constraints enforce that $z$ is a valid portfolio. In practice, the deterministic solution may concentrate heavily on a few assets and yield a low average return, whereas a stochastic approach can achieve higher returns by accounting for variance.

\textbf{Experimental setup.} We have daily prices and volumes spanning 2004-2017 and evaluate on the S\&P 500 index constituents~\citep{quandl_wiki_dataset_nasdaq_2025}. The features $x \in \mathbb{R}^{28}$ include recent historical return, trading volume windows, and rolling averages. The immediate-return predictor $P_\theta(\cdot|x)$ is to predict the next day's price. 
% Data are split chronologically 70/10/20 (train/val/test). 
We report the performance of different DFL baselines with 50 portfolios in~\Cref{tab:exp_all_methods} and other sizes of portfolios in~\Cref{sec:ablation_study}. 
%Our diffusion DFL yields the lowest regret across all sizes, outperforming both deterministic and Gaussian DFL baselines.

\begin{table}[htbp]
% \vspace{-0.1cm}
\centering
\caption{Results for different optimization tasks. Our two diffusion DFL methods achieve the best and second-best decision quality in all 3 tasks, significantly better than other baselines. \textbf{Bolded} values are the best in test task losses; \second{underlined} values are the 2nd-best. Mean $\pm$ standard error across 10 runs. }
\vspace{0.2cm}
\label{tab:exp_all_methods}
\setlength{\tabcolsep}{2pt}
\footnotesize
\begin{tabular}{l cc cc cc}
\toprule
 & \multicolumn{2}{c}{\textbf{Synthetic Example}} & \multicolumn{2}{c}{\textbf{Power Schedule}} & \multicolumn{2}{c}{\textbf{Stock Portfolio}} \\
\cmidrule(lr){2-3}\cmidrule(lr){4-5}\cmidrule(lr){6-7}
\textbf{Label / Method} & \textbf{RMSE}$\downarrow$ & \textbf{Task}$\downarrow$ & \textbf{RMSE}$\downarrow$ & \textbf{Task}$\downarrow$ & \textbf{RMSE}$\downarrow$ & \textbf{Task (\%)}$\uparrow$ \\
\midrule
\multicolumn{7}{l}{\emph{Two-stage (TS)}}\\
\rowcolor{black!6} Deterministic TS & {0.639}$_{\textcolor{gray}{\pm0.00}}$ & 1.987$_{\textcolor{gray}{\pm0.00}}$ & {0.120}$_{\textcolor{gray}{\pm0.00}}$ & 41.239$_{\textcolor{gray}{\pm3.18}}$ & {0.027}$_{\textcolor{gray}{\pm0.00}}$ & 0.04\%$_{\textcolor{gray}{\pm0.04}}$ \\
\rowcolor{black!6} Gaussian TS      & 0.720$_{\textcolor{gray}{\pm0.00}}$ & 1.272$_{\textcolor{gray}{\pm0.23}}$ & {0.117}$_{\textcolor{gray}{\pm0.00}}$ & 5.580$_{\textcolor{gray}{\pm0.45}}$ & 0.188$_{\textcolor{gray}{\pm0.03}}$ & 0.10\%$_{\textcolor{gray}{\pm0.04}}$ \\
\rowcolor{black!6} Diffusion TS     & 0.905$_{\textcolor{gray}{\pm0.00}}$ & {0.393}$_{\textcolor{gray}{\pm0.00}}$ & 0.147$_{\textcolor{gray}{\pm0.00}}$ & 7.901$_{\textcolor{gray}{\pm0.76}}$  & 0.455$_{\textcolor{gray}{\pm0.00}}$ & 0.13\%$_{\textcolor{gray}{\pm0.03}}$ \\
\addlinespace[2pt]
\multicolumn{7}{l}{\emph{Decision-focused learning (DFL)}}\\
Deterministic        & {0.640}$_{\textcolor{gray}{\pm0.00}}$ & 1.987$_{\textcolor{gray}{\pm0.00}}$ & 4.997$_{\textcolor{gray}{\pm0.10}}$ & 4.324$_{\textcolor{gray}{\pm0.25}}$ & {0.032}$_{\textcolor{gray}{\pm0.00}}$ & 0.07\%$_{\textcolor{gray}{\pm0.00}}$ \\
Gaussian Reparameterization      & 0.707$_{\textcolor{gray}{\pm0.00}}$ & 1.169$_{\textcolor{gray}{\pm0.03}}$ & 4.525$_{\textcolor{gray}{\pm0.12}}$ & 3.724$_{\textcolor{gray}{\pm0.05}}$ & 0.189$_{\textcolor{gray}{\pm0.03}}$ & 0.08\%$_{\textcolor{gray}{\pm0.03}}$ \\
Gaussian Score Function    & 0.708$_{\textcolor{gray}{\pm0.00}}$ & 1.132$_{\textcolor{gray}{\pm0.00}}$ & 4.713$_{\textcolor{gray}{\pm0.15}}$ & 4.087$_{\textcolor{gray}{\pm0.06}}$ & 0.187$_{\textcolor{gray}{\pm0.03}}$ & 0.14\%$_{\textcolor{gray}{\pm0.05}}$ \\
Diffusion Reparameterization    & 0.852$_{\textcolor{gray}{\pm0.01}}$ & \second{0.365}$_{\textcolor{gray}{\pm0.00}}$ & \,3.141$_{\textcolor{gray}{\pm0.06}}$ & \textbf{3.152}$_{\textcolor{gray}{\pm0.03}}$ & 0.063$_{\textcolor{gray}{\pm0.00}}$ & \textbf{4.17\%}$_{\textcolor{gray}{\pm0.24}}$ \\
Diffusion Score Function   & 0.849$_{\textcolor{gray}{\pm0.09}}$ & \best{0.362}$_{\textcolor{gray}{\pm0.00}}$ & \,2.893$_{\textcolor{gray}{\pm0.03}}$ & \second{3.171}$_{\textcolor{gray}{\pm0.02}}$ & 0.067$_{\textcolor{gray}{\pm0.00}}$ & \second{3.98\%}$_{\textcolor{gray}{\pm0.31}}$ \\
\bottomrule
\end{tabular}
\end{table}

\section{Discussion of Experimental Results and Ablation Study}

\subsection{Discussion of Results in~\Cref{tab:exp_all_methods}}
\paragraph{Two-stage vs DFL.} As shown in \Cref{tab:exp_all_methods}, across all three experiment tasks, we find that end-to-end DFL leads to better downstream decisions than the conventional two-stage approach. Conventional two-stage methods minimize RMSE during training, but this often leads to poor downstream decisions. In contrast, all variants of DFL directly minimize the decision cost during training and thus achieve lower decision costs.

\paragraph{Deterministic vs Stochastic Optimization.}
Our results show that stochastic DFL methods outperform deterministic DFL in terms of decision quality on every task. By modeling uncertainty, stochastic predictors enable the decision optimization to account for risk and variability in outcomes. For instance, in the portfolio experiment, the deterministic DFL yields only $0.07\%$ return, whereas a Gaussian DFL modestly improves that, and our diffusion DFL achieves nearly $4\%$ average return. These gains come from the stochastic models' ability to predict uncertainty: instead of committing to a point prediction of $y$, the stochastic DFL produces decisions for a range of possible outcomes.

\paragraph{Benefits of Diffusion DFL.} Among the stochastic approaches, including baselines using Gaussian models, our diffusion DFL method consistently delivers the best decision performance. In particular, the diffusion model's strength is the capacity to capture complex, multi-modal outcome distributions that a simple parametric Gaussian cannot represent. The Gaussian DFL sometimes falls short of the optimal decision quality. The diffusion model, on the other hand, can represent more intricate distributions of $y$, leading to decisions that better reflect complex scenarios.

\subsection{Ablation Study}\label{sec:ablation_study}
% In this section, we analyze two important training components in our diffusion DFL method: the gradient estimation techniques (reparameterization vs. score-function) and the effect of using a small subset of diffusion timesteps during training over the real-world power schedule task.

\textbf{Comparison Cost for Reparameterization and Score function.}
A key finding from our ablation study is the computational advantage of score-function approach over the reparameterization. Here, we measure the trade-off between training cost and the final decision performance for different gradient estimators and sampling budgets.

\begin{figure}[htbp]
  \centering
  \vspace{-0.2cm}
  \begin{minipage}[t]{0.66\linewidth}
    \centering
    \includegraphics[width=\linewidth]{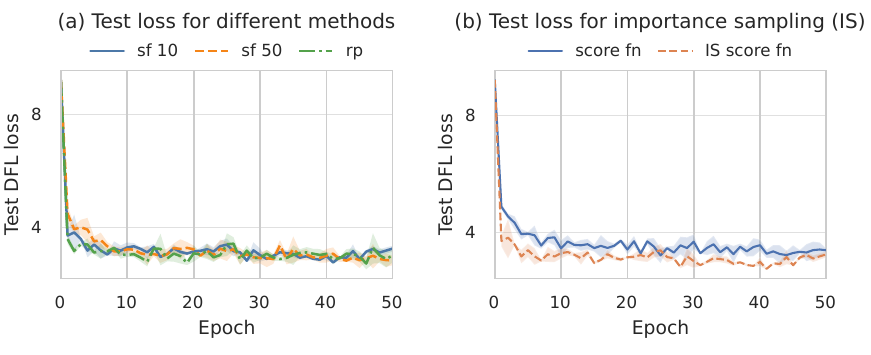}
    \vspace{-0.8cm}
    \caption{Learning curves for (a) score function with 10 and 50 samples (\textit{sf 10} and \textit{sf 50}) and reparameterization (\textit{rp}), (b) score function and importance-weighted score function with 10 samples.}
    \label{fig:learning_curve}
  \end{minipage}\hfill
  \begin{minipage}[t]{0.313\linewidth}
    \centering
    \includegraphics[width=\linewidth]{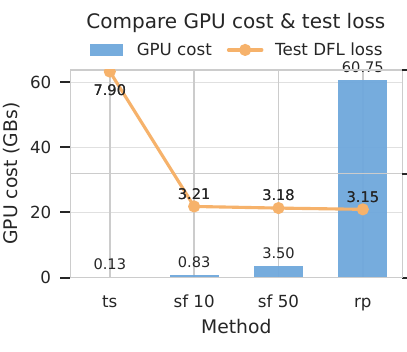}
    \vspace{-0.8cm}
    \caption{Computation cost vs. performance trade-off for diffusion DFL training}
    \label{fig:abl_gpu}
  \end{minipage}
\end{figure}

In \Cref{fig:learning_curve} (a), we see that all variants reach similar final performance on the test set, indicating that even using as few as 50 samples is sufficient to optimize the decision quality accurately. 
\Cref{fig:abl_gpu} plots the GPU memory cost alongside the final test loss. The reparameterization method is very computationally expensive, requiring about 60 GB of GPU memory for backpropagating through all diffusion steps. In contrast, the score-function with 50 samples achieves virtually the same test loss as the reparameterization method (difference within 0.02) while using an order of magnitude less memory. Even with 10 samples, though slightly worse in loss, it still outperforms the deterministic baseline and uses a tiny fraction of the compute. These results validate that the score-function approach retains the decision-quality benefits of diffusion DFL while dramatically cutting computational requirements, making diffusion DFL practical even for complex problems.

\textbf{Gradient variance reduction.}
As discussed in~\Cref{sec:discusion_sf}, using the score function estimator allows us to avoid backpropagating through the entire diffusion sampling process by only sampling a limited number of diffusion timesteps per update. The reason behind this is that a naive implementation, sampling timesteps uniformly at random, would yield a very high variance in the gradient estimates, which then leads to unstable training. Intuitively, early diffusion steps (large noise levels) dominate the ELBO loss and its gradients, so if they happen to be sampled, they contribute disproportionately and noisily. With a small random subset of timesteps, the gradient estimate can thus be highly imbalanced and noisy, which causes training divergence in practice.

To address this, we adopt an importance sampling strategy for choosing diffusion timesteps. Empirically, as shown in~\Cref{fig:learning_curve} (b), the learning curves with the importance-weighted sampler are much smoother and more stable than with the uniform sampler. The score-function DFL training no longer diverges; instead, it converges cleanly, indicating that our variance reduction strategy successfully stabilizes the training process for diffusion DFL.

\begin{wrapfigure}{r}{0.5\textwidth}
  \vspace{-0.2cm}
  \centering
  \includegraphics[width=\linewidth]{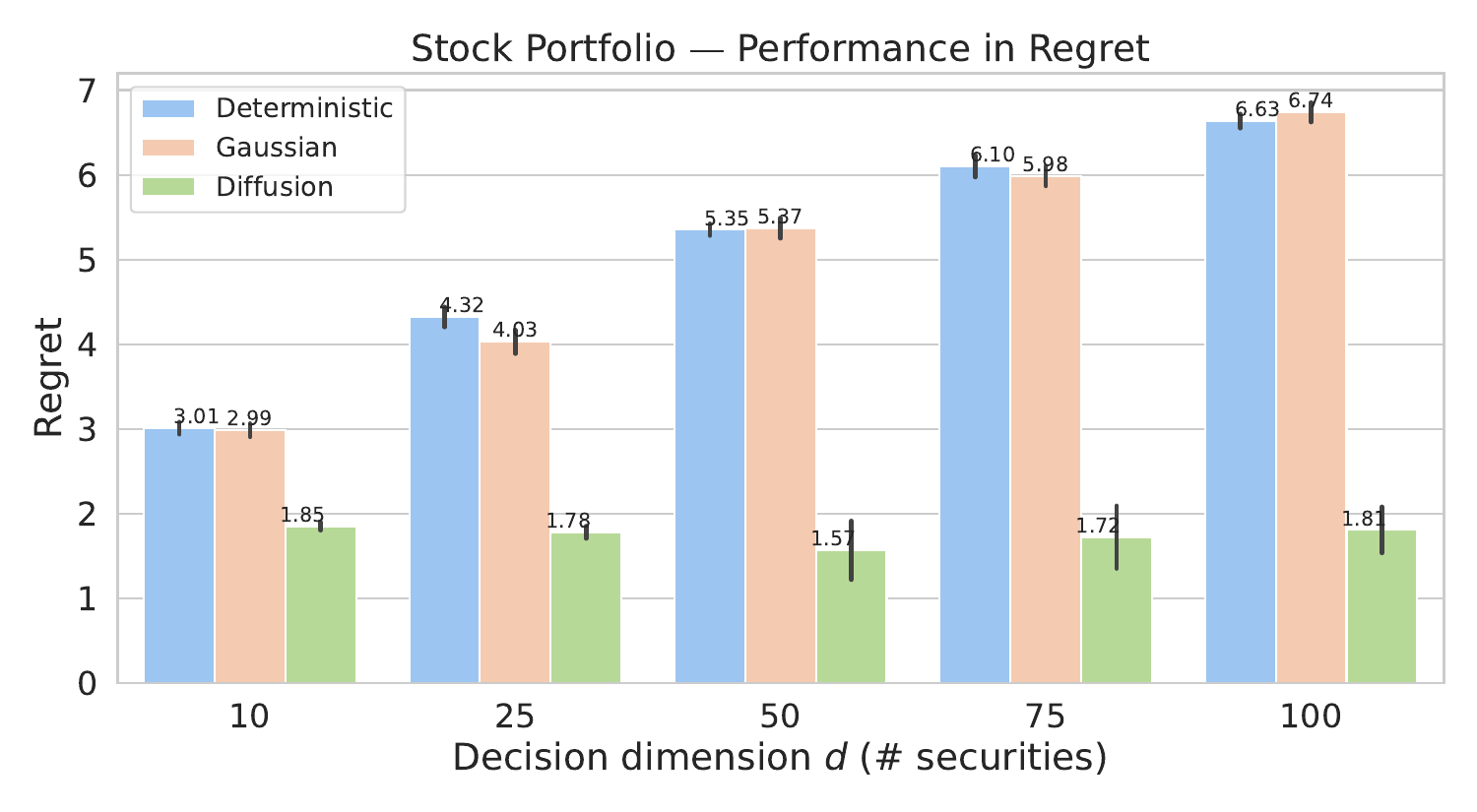}
  \vspace{-0.8cm}
  \caption{Test regret vs. decision dimension $d$ in the stock portfolio task.}
  \label{fig:portfolio_regret}
  \vspace{-6pt}
\end{wrapfigure}
\textbf{Comparison on different problem sizes.}
A key challenge for DFL is scalability: as the decision dimension grows, many methods degrade significantly~\cite{mandi_decision-focused_2024}. In this experiment, we investigate the performance of DFL methods under various decision dimensions in the stock portfolio. Specifically, we set the decision dimension range from 10 to 100 and report the test regrets. As summarized in \Cref{fig:portfolio_regret}, the regret gap between DFL diffusion and Gaussian and deterministic methods increases with increasing dimension, which demonstrates that DFL diffusion scales effectively in more complex decision settings.
\section{Conclusion}
We propose the first diffusion-based DFL approach for stochastic optimization, which trains a diffusion model to capture complex uncertainty in problem parameters. We develop two end-to-end training techniques to integrate the diffusion model into decision-making: reparameterization and score function approximation. As demonstrated with empirical evidence, the score function method drastically reduces memory and computation cost while achieving similar performance to reparameterization and being easy to train. Empirically, diffusion DFL achieves state-of-the-art results on multiple benchmarks, consistently outperforming both traditional two-stage methods and prior DFL approaches.
\section*{Reproducibility statement}
We release an anonymized repository containing all code, configuration files, and scripts needed to reproduce our results, including data generation and figure plotting. All proofs for the main paper are stated in the appendix with explanations and proper assumptions. 

\bibliographystyle{plainnat}
\bibliography{references_zihao}

\begin{thebibliography}{36}
\providecommand{\natexlab}[1]{#1}
\providecommand{\url}[1]{\texttt{#1}}
\expandafter\ifx\csname urlstyle\endcsname\relax
  \providecommand{\doi}[1]{doi: #1}\else
  \providecommand{\doi}{doi: \begingroup \urlstyle{rm}\Url}\fi

\bibitem[Amos and Kolter(2017)]{amos_optnet_2017}
Brandon Amos and J.~Zico Kolter.
\newblock {OptNet}: {Differentiable} {Optimization} as a {Layer} in {Neural} {Networks}.
\newblock In \emph{Proceedings of the 34th {International} {Conference} on {Machine} {Learning}}, pages 136--145. PMLR, July 2017.
\newblock URL \url{https://proceedings.mlr.press/v70/amos17a.html}.
\newblock ISSN: 2640-3498.

\bibitem[Arjevani et~al.(2020)Arjevani, Carmon, Duchi, Foster, Sekhari, and Sridharan]{arjevani_second-order_2020}
Yossi Arjevani, Yair Carmon, John~C. Duchi, Dylan~J. Foster, Ayush Sekhari, and Karthik Sridharan.
\newblock Second-{Order} {Information} in {Non}-{Convex} {Stochastic} {Optimization}: {Power} and {Limitations}.
\newblock In \emph{Proceedings of {Twenty} {Second} {Conference} on {Learning} {Theory}}, pages 242--299. PMLR, July 2020.
\newblock URL \url{https://proceedings.mlr.press/v125/arjevani20a.html}.
\newblock ISSN: 2640-3498.

\bibitem[Bertsimas and Kallus(2020)]{bertsimas_predictive_2020}
Dimitris Bertsimas and Nathan Kallus.
\newblock From {Predictive} to {Prescriptive} {Analytics}.
\newblock \emph{Management Science}, 66\penalty0 (3):\penalty0 1025--1044, March 2020.
\newblock ISSN 0025-1909, 1526-5501.
\newblock \doi{10.1287/mnsc.2018.3253}.
\newblock URL \url{https://pubsonline.informs.org/doi/10.1287/mnsc.2018.3253}.

\bibitem[Dhariwal and Nichol(2021)]{dhariwal_diffusion_2021}
Prafulla Dhariwal and Alexander Nichol.
\newblock Diffusion {Models} {Beat} {GANs} on {Image} {Synthesis}.
\newblock In \emph{Advances in {Neural} {Information} {Processing} {Systems}}, volume~34, pages 8780--8794. Curran Associates, Inc., 2021.
\newblock URL \url{https://proceedings.neurips.cc/paper_files/paper/2021/hash/49ad23d1ec9fa4bd8d77d02681df5cfa-Abstract.html}.

\bibitem[Donti et~al.(2017)Donti, Amos, and Kolter]{donti_task-based_2017}
Priya~L. Donti, Brandon Amos, and J.~Zico Kolter.
\newblock Task-based {End}-to-end {Model} {Learning} in {Stochastic} {Optimization}.
\newblock In \emph{Advances in {Neural} {Information} {Processing} {Systems}}, volume~30, Long Beach, CA, USA, December 2017. Curran Associates, Inc.
\newblock URL \url{http://papers.nips.cc/paper/7132-task-based-end-to-end-model-learning-in-stochastic-optimization}.

\bibitem[Elmachtoub and Grigas(2022)]{elmachtoub_smart_2022}
Adam~N. Elmachtoub and Paul Grigas.
\newblock Smart “{Predict}, then {Optimize}”.
\newblock \emph{Management Science}, 68\penalty0 (1):\penalty0 9--26, January 2022.
\newblock ISSN 0025-1909.
\newblock \doi{10.1287/mnsc.2020.3922}.
\newblock URL \url{https://pubsonline.informs.org/doi/10.1287/mnsc.2020.3922}.
\newblock Publisher: INFORMS.

\bibitem[Ho et~al.(2020)Ho, Jain, and Abbeel]{ho_denoising_2020}
Jonathan Ho, Ajay Jain, and Pieter Abbeel.
\newblock Denoising {Diffusion} {Probabilistic} {Models}.
\newblock In \emph{Advances in {Neural} {Information} {Processing} {Systems}}, volume~33, pages 6840--6851. Curran Associates, Inc., 2020.
\newblock URL \url{https://proceedings.neurips.cc/paper/2020/hash/4c5bcfec8584af0d967f1ab10179ca4b-Abstract.html}.

\bibitem[Hoffman and Johnson(2016)]{hoffman_elbo_2016}
Matthew~D Hoffman and Matthew~J Johnson.
\newblock {ELBO} surgery: yet another way to carve up the variational evidence lower bound.
\newblock In \emph{{NIPS} 2016 workshop}, 2016.

\bibitem[Jeon et~al.(2025)Jeon, Bae, Park, Kim, and Kim]{jeon_locally_2025}
Haeun Jeon, Hyunglip Bae, Minsu Park, Chanyeong Kim, and Woo~Chang Kim.
\newblock Locally {Convex} {Global} {Loss} {Network} for {Decision}-{Focused} {Learning}.
\newblock \emph{Proceedings of the AAAI Conference on Artificial Intelligence}, 39\penalty0 (25):\penalty0 26805--26812, April 2025.
\newblock ISSN 2374-3468.
\newblock \doi{10.1609/aaai.v39i25.34884}.
\newblock URL \url{https://ojs.aaai.org/index.php/AAAI/article/view/34884}.

\bibitem[Kim et~al.(2015)Kim, Pasupathy, and Henderson]{kim_guide_2015}
Sujin Kim, Raghu Pasupathy, and Shane~G. Henderson.
\newblock A {Guide} to {Sample} {Average} {Approximation}.
\newblock In Michael~C Fu, editor, \emph{Handbook of {Simulation} {Optimization}}, pages 207--243. Springer, New York, NY, 2015.
\newblock ISBN 978-1-4939-1384-8.
\newblock \doi{10.1007/978-1-4939-1384-8_8}.
\newblock URL \url{https://doi.org/10.1007/978-1-4939-1384-8_8}.

\bibitem[Kingma and Welling(2014)]{Kingma2014}
Diederik~P Kingma and Max Welling.
\newblock Auto-{Encoding} {Variational} {Bayes}.
\newblock In \emph{International {Conference} on {Learning} {Representations}}, April 2014.
\newblock URL \url{http://arxiv.org/abs/1312.6114}.

\bibitem[Kleywegt et~al.(2002)Kleywegt, Shapiro, and Homem-de Mello]{kleywegt_sample_2002}
Anton~J. Kleywegt, Alexander Shapiro, and Tito Homem-de Mello.
\newblock The {Sample} {Average} {Approximation} {Method} for {Stochastic} {Discrete} {Optimization}.
\newblock \emph{SIAM Journal on Optimization}, 12\penalty0 (2):\penalty0 479--502, January 2002.
\newblock ISSN 1052-6234, 1095-7189.
\newblock \doi{10.1137/S1052623499363220}.
\newblock URL \url{http://epubs.siam.org/doi/10.1137/S1052623499363220}.

\bibitem[Kochenderfer et~al.(2015)Kochenderfer, Amato, Chowdhary, How, Reynolds, Thornton, Torres-Carrasquillo, Üre, and Vian]{kochenderfer_decision_2015}
Mykel~J. Kochenderfer, Christopher Amato, Girish Chowdhary, Jonathan~P. How, Hayley J.~Davison Reynolds, Jason~R. Thornton, Pedro~A. Torres-Carrasquillo, N.~Kemal Üre, and John Vian.
\newblock \emph{Decision {Making} {Under} {Uncertainty}: {Theory} and {Application}}.
\newblock The MIT Press, 1st edition, June 2015.
\newblock ISBN 978-0-262-02925-4.

\bibitem[Kong et~al.(2025)Kong, Du, Mu, Neklyudov, Bortoli, Wu, Wang, Ferber, Ma, Gomes, and Zhang]{kong_diffusion_2025}
Lingkai Kong, Yuanqi Du, Wenhao Mu, Kirill Neklyudov, Valentin~De Bortoli, Dongxia Wu, Haorui Wang, Aaron~M. Ferber, Yian Ma, Carla~P. Gomes, and Chao Zhang.
\newblock Diffusion {Models} as {Constrained} {Samplers} for {Optimization} with {Unknown} {Constraints}.
\newblock In \emph{Proceedings of {The} 28th {International} {Conference} on {Artificial} {Intelligence} and {Statistics}}, pages 4582--4590. PMLR, April 2025.
\newblock URL \url{https://proceedings.mlr.press/v258/kong25b.html}.
\newblock ISSN: 2640-3498.

\bibitem[Krishnamoorthy et~al.(2023)Krishnamoorthy, Mashkaria, and Grover]{krishnamoorthy_diffusion_2023}
Siddarth Krishnamoorthy, Satvik~Mehul Mashkaria, and Aditya Grover.
\newblock Diffusion {Models} for {Black}-{Box} {Optimization}.
\newblock In \emph{Proceedings of the 40th {International} {Conference} on {Machine} {Learning}}, pages 17842--17857. PMLR, July 2023.
\newblock URL \url{https://proceedings.mlr.press/v202/krishnamoorthy23a.html}.
\newblock ISSN: 2640-3498.

\bibitem[Mandi et~al.(2022)Mandi, Bucarey, Tchomba, and Guns]{mandi_decision-focused_2022}
Jayanta Mandi, Victor Bucarey, Maxime Mulamba~Ke Tchomba, and Tias Guns.
\newblock Decision-{Focused} {Learning}: {Through} the {Lens} of {Learning} to {Rank}.
\newblock In \emph{Proceedings of the 39th {International} {Conference} on {Machine} {Learning}}, pages 14935--14947. PMLR, June 2022.
\newblock URL \url{https://proceedings.mlr.press/v162/mandi22a.html}.
\newblock ISSN: 2640-3498.

\bibitem[Mandi et~al.(2024)Mandi, Kotary, Berden, Mulamba, Bucarey, Guns, and Fioretto]{mandi_decision-focused_2024}
Jayanta Mandi, James Kotary, Senne Berden, Maxime Mulamba, Victor Bucarey, Tias Guns, and Ferdinando Fioretto.
\newblock Decision-{Focused} {Learning}: {Foundations}, {State} of the {Art}, {Benchmark} and {Future} {Opportunities}.
\newblock \emph{Journal of Artificial Intelligence Research}, 80:\penalty0 1623--1701, August 2024.
\newblock ISSN 1076-9757.
\newblock \doi{10.1613/jair.1.15320}.
\newblock URL \url{https://www.jair.org/index.php/jair/article/view/15320}.

\bibitem[Mohamed et~al.(2020)Mohamed, Rosca, Figurnov, and Mnih]{mohamed_monte_2020}
Shakir Mohamed, Mihaela Rosca, Michael Figurnov, and Andriy Mnih.
\newblock Monte {Carlo} {Gradient} {Estimation} in {Machine} {Learning}.
\newblock \emph{Journal of Machine Learning Research}, 21\penalty0 (132):\penalty0 1--62, 2020.
\newblock ISSN 1533-7928.
\newblock URL \url{http://jmlr.org/papers/v21/19-346.html}.

\bibitem[Nichol and Dhariwal(2021)]{nichol_improved_2021}
Alexander~Quinn Nichol and Prafulla Dhariwal.
\newblock Improved {Denoising} {Diffusion} {Probabilistic} {Models}.
\newblock In \emph{Proceedings of the 38th {International} {Conference} on {Machine} {Learning}}, pages 8162--8171. PMLR, July 2021.
\newblock URL \url{https://proceedings.mlr.press/v139/nichol21a.html}.
\newblock ISSN: 2640-3498.

\bibitem[{PJM Interconnection}(2025)]{pjm_interconnection_data_2025}
{PJM Interconnection}.
\newblock Data {Miner}, 2025.
\newblock URL \url{https://dataminer2.pjm.com/list}.

\bibitem[{Quandl WIKI dataset}(2025)]{quandl_wiki_dataset_nasdaq_2025}
{Quandl WIKI dataset}.
\newblock Nasdaq {Data} {Link}, 2025.
\newblock URL \url{https://data.nasdaq.com}.

\bibitem[Sanokowski et~al.(2025)Sanokowski, Hochreiter, and Lehner]{sanokowski_diffusion_2025}
Sebastian Sanokowski, Sepp Hochreiter, and Sebastian Lehner.
\newblock A {Diffusion} {Model} {Framework} for {Unsupervised} {Neural} {Combinatorial} {Optimization}, August 2025.
\newblock URL \url{http://arxiv.org/abs/2406.01661}.
\newblock arXiv:2406.01661 [cs].

\bibitem[Shah et~al.(2022)Shah, Wang, Wilder, Perrault, and Tambe]{shah_decision-focused_2022}
Sanket Shah, Kai Wang, Bryan Wilder, Andrew Perrault, and Milind Tambe.
\newblock Decision-{Focused} {Learning} without {Decision}-{Making}: {Learning} {Locally} {Optimized} {Decision} {Losses}.
\newblock \emph{Advances in Neural Information Processing Systems}, 35:\penalty0 1320--1332, December 2022.
\newblock URL \url{https://proceedings.neurips.cc/paper_files/paper/2022/hash/0904c7edde20d7134a77fc7f9cd86ea2-Abstract-Conference.html}.

\bibitem[Shalev-Shwartz et~al.(2009)Shalev-Shwartz, Shamir, Srebro, and Sridharan]{shalev-shwartz_stochastic_2009}
Shai Shalev-Shwartz, Ohad Shamir, Nathan Srebro, and Karthik Sridharan.
\newblock Stochastic {Convex} {Optimization}.
\newblock In \emph{Proceedings of {Thirty} {Third} {Conference} on {Learning} {Theory}}. PMLR, 2009.

\bibitem[Shariatmadar et~al.(2025)Shariatmadar, Yorke-Smith, Osman, Cuzzolin, Hallez, and Moens]{shariatmadar_generalized_2025}
Keivan Shariatmadar, Neil Yorke-Smith, Ahmad Osman, Fabio Cuzzolin, Hans Hallez, and David Moens.
\newblock Generalized {Decision} {Focused} {Learning} under {Imprecise} {Uncertainty}--{Theoretical} {Study}, March 2025.
\newblock URL \url{http://arxiv.org/abs/2502.17984}.
\newblock arXiv:2502.17984 [cs].

\bibitem[Silvestri et~al.(2023)Silvestri, Berden, Mandi, Mahmutoğulları, Mulamba, Filippo, Guns, and Lombardi]{silvestri_score_2023}
Mattia Silvestri, Senne Berden, Jayanta Mandi, Ali~İrfan Mahmutoğulları, Maxime Mulamba, Allegra~De Filippo, Tias Guns, and Michele Lombardi.
\newblock Score {Function} {Gradient} {Estimation} to {Widen} the {Applicability} of {Decision}-{Focused} {Learning}.
\newblock September 2023.
\newblock URL \url{https://openreview.net/forum?id=ty046JUllZ}.

\bibitem[Sohl-Dickstein et~al.(2015)Sohl-Dickstein, Weiss, Maheswaranathan, and Ganguli]{sohl-dickstein_deep_2015}
Jascha Sohl-Dickstein, Eric Weiss, Niru Maheswaranathan, and Surya Ganguli.
\newblock Deep {Unsupervised} {Learning} using {Nonequilibrium} {Thermodynamics}.
\newblock In \emph{Proceedings of the 32nd {International} {Conference} on {Machine} {Learning}}, pages 2256--2265. PMLR, June 2015.
\newblock URL \url{https://proceedings.mlr.press/v37/sohl-dickstein15.html}.
\newblock ISSN: 1938-7228.

\bibitem[Song and Ermon(2019)]{song_generative_2019}
Yang Song and Stefano Ermon.
\newblock Generative {Modeling} by {Estimating} {Gradients} of the {Data} {Distribution}.
\newblock In \emph{Advances in {Neural} {Information} {Processing} {Systems}}, volume~32. Curran Associates, Inc., 2019.
\newblock URL \url{https://proceedings.neurips.cc/paper_files/paper/2019/hash/3001ef257407d5a371a96dcd947c7d93-Abstract.html}.

\bibitem[Song et~al.(2020)Song, Sohl-Dickstein, Kingma, Kumar, Ermon, and Poole]{song_score-based_2020}
Yang Song, Jascha Sohl-Dickstein, Diederik~P. Kingma, Abhishek Kumar, Stefano Ermon, and Ben Poole.
\newblock Score-{Based} {Generative} {Modeling} through {Stochastic} {Differential} {Equations}.
\newblock October 2020.
\newblock URL \url{https://openreview.net/forum?id=PxTIG12RRHS&utm_campaign=NLP%20News&utm_medium=email&utm_source=Revue%20newsletter}.

\bibitem[Sun and Yang(2023)]{sun_difusco_2023}
Zhiqing Sun and Yiming Yang.
\newblock {DIFUSCO}: {Graph}-based {Diffusion} {Solvers} for {Combinatorial} {Optimization}.
\newblock November 2023.
\newblock URL \url{https://openreview.net/forum?id=JV8Ff0lgVV}.

\bibitem[Tang and Nurmaya~Musa(2011)]{tang_identifying_2011}
Ou~Tang and S.~Nurmaya~Musa.
\newblock Identifying risk issues and research advancements in supply chain risk management.
\newblock \emph{International Journal of Production Economics}, 133\penalty0 (1):\penalty0 25--34, September 2011.
\newblock ISSN 0925-5273.
\newblock \doi{10.1016/j.ijpe.2010.06.013}.
\newblock URL \url{https://www.sciencedirect.com/science/article/pii/S0925527310002215}.

\bibitem[Tashiro et~al.(2021)Tashiro, Song, Song, and Ermon]{tashiro_csdi_2021}
Yusuke Tashiro, Jiaming Song, Yang Song, and Stefano Ermon.
\newblock {CSDI}: {Conditional} {Score}-based {Diffusion} {Models} for {Probabilistic} {Time} {Series} {Imputation}.
\newblock In \emph{Advances in {Neural} {Information} {Processing} {Systems}}, volume~34, pages 24804--24816. Curran Associates, Inc., 2021.
\newblock URL \url{https://proceedings.neurips.cc/paper_files/paper/2021/hash/cfe8504bda37b575c70ee1a8276f3486-Abstract.html}.

\bibitem[Vahdat and Kautz(2020)]{vahdat_nvae_2020}
Arash Vahdat and Jan Kautz.
\newblock {NVAE}: {A} {Deep} {Hierarchical} {Variational} {Autoencoder}.
\newblock In \emph{Advances in {Neural} {Information} {Processing} {Systems}}, volume~33, pages 19667--19679. Curran Associates, Inc., 2020.
\newblock URL \url{https://proceedings.neurips.cc/paper/2020/hash/e3b21256183cf7c2c7a66be163579d37-Abstract.html}.

\bibitem[Wang et~al.(2025)Wang, Liang, Chen, Fioretto, and Zhu]{wang_gen-dfl_2025}
Prince~Zizhuang Wang, Jinhao Liang, Shuyi Chen, Ferdinando Fioretto, and Shixiang Zhu.
\newblock Gen-{DFL}: {Decision}-{Focused} {Generative} {Learning} for {Robust} {Decision} {Making}, February 2025.
\newblock URL \url{http://arxiv.org/abs/2502.05468}.
\newblock arXiv:2502.05468 [cs].

\bibitem[Wang et~al.(2024)Wang, Pan, Li, Lu, Kong, Jiang, and Kong]{wang_sample_2024}
Yafei Wang, Bo~Pan, Mei Li, Jianya Lu, Lingchen Kong, Bei Jiang, and Linglong Kong.
\newblock Sample {Average} {Approximation} for {Conditional} {Stochastic} {Optimization} with {Dependent} {Data}.
\newblock June 2024.
\newblock URL \url{https://openreview.net/forum?id=YuGnRORkJm}.

\bibitem[Wilder et~al.(2019)Wilder, Dilkina, and Tambe]{wilder_melding_2019}
Bryan Wilder, Bistra Dilkina, and Milind Tambe.
\newblock Melding the {Data}-{Decisions} {Pipeline}: {Decision}-{Focused} {Learning} for {Combinatorial} {Optimization}.
\newblock \emph{Proceedings of the AAAI Conference on Artificial Intelligence}, 33\penalty0 (01):\penalty0 1658--1665, July 2019.
\newblock ISSN 2374-3468.
\newblock \doi{10.1609/aaai.v33i01.33011658}.
\newblock URL \url{https://ojs.aaai.org/index.php/AAAI/article/view/3982}.

\end{thebibliography}

\newpage
\onecolumn
\appendix
\appendix
\newpage
\section{Appendix}
\paragraph{Notation.}
We let $\pderiv{f}{x}$ denote the Jacobian matrix where $\big(\pderiv{f}{x}\big)_{i,j} := \pderiv{f_i}{x_j}$ and $\nabla_x f := \big(\pderiv{f}{x}\big)^\top$ denote the gradient. For a vector $v$, $D(v)$ denotes a diagonal matrix with $v$ on its diagonal. Let $P(\cdot)$ denote a probability distribution and $p(\cdot)$ denote a probability density; in particular, for diffusion models we use $P_\theta$ for the model's output distribution and $p_\theta$ for transition densities. % We use $[T] := \{1, \dotsc, T\}$. 

In this appendix, we derive the decision optimization problem with \textbf{general convex constraints} rather than merely linear constraints. Assume the optimization problem is
\begin{equation}
    z_\theta^*(x) = \argmin_z \E_{y \sim P_\theta(\cdot|x)} [f(y,z)], \quad\text{s.t.} \ \ h(x, z) \leq 0,\ g(x, z) = 0,
\end{equation}
where $h(x, z) \leq 0$ denotes the convex inequalities constraints and $g(x, z) = 0$ denotes the equality constraints.

\subsection{Proofs for Section~\ref{sec:reparm}}\label{apd:proof_rp}
% \begin{proposition}
% Assume the model prediction $y$ can be expressed as a transformation $y = T(\epsilon, \theta \mid x)$ of a base noise distribution $\epsilon \sim P(\epsilon)$, where $T$ is differentiable in $\theta$. Also assume $\LL(x, z;\theta)$ is continuously differentiable. Then we have
% \begin{align}
%      \nabla_\theta{\LL(x, z;\theta)} = \E_{\epsilon \sim P(\epsilon)} [(\nabla_\theta T(\epsilon, \theta|x))^\top \nabla_y f(y,z)].
% \end{align}
% \end{proposition}

% \begin{proof}
%     \begin{align}
%      \nabla_\theta{\LL(x, z;\theta)} &= \nabla_\theta \E_{y \sim P_\theta(\cdot\mid x)} [f(y,z)] \\
%      &= \nabla_\theta \E_{\epsilon \sim P(\epsilon)}[f(T(\epsilon, \theta|x), z)] \\
%      &= \E_{\epsilon \sim P(\epsilon)}[\nabla_\theta f(T(\epsilon, \theta|x), z)] \\
%      &= \E_{\epsilon \sim P(\epsilon)}\left[ \nabla_\theta T(\epsilon, \theta \mid x)^\top \nabla_y f(T(\epsilon, \theta|x),z) \right]
% \end{align}
% \end{proof}

\begin{proposition}[Reparameterization trick in diffusion models]\label{prop:reparam}
Let $T \in \N^+$, and suppose the reverse diffusion model defines a Gaussian distribution in Eq.~\ref{eq:diffusion_reverse} with fixed scalars $\sigma_t \geq 0$ and a standard normal prior $y_T \sim \calN(0, I)$. Let $\{\epsilon_t\}_{t=0}^T$ be i.i.d. $\calN(0, I)$. Then the model output $y$ can be expressed as a transformation $y = R(\epsilon_{0:T}, \theta \mid x)$ of a base noise distribution $\epsilon \sim P(\epsilon)$, where $R$ is differentiable in $\theta$. Also assume $\E_{y \sim P_\theta(\cdot|x)} [f(y,z)]$ is continuously differentiable. Then we have
\begin{align}
     \nabla_\theta{\E_{y \sim P_\theta(\cdot|x)} [f(y,z)]} = \E_{\epsilon \sim P(\epsilon)}\left[ \left(\sum_{s=1}^T \left(\prod_{u=1}^{s-1} J_u\right) A_s\right)^\top \nabla_y f(R(\epsilon, \theta \mid x),z) \right],
\end{align}
where $A_t := \frac{\partial \mu_\theta (y_t, t, x)}{\partial \theta}$, $J_t := \frac{\partial \mu_\theta(y_t, t, x)}{\partial y_t}$, and we define $\prod_{u=1}^0 J_u := I$.
\end{proposition}

\begin{proof}
The conditional diffusion reverse process is defined as
\[
    y_{t-1} = \mu_\theta (y_t, t, x) + \sigma_t \epsilon_{t-1},
    \qquad
    y_T = \epsilon_T,
\]
where the noise term $\sigma_t \epsilon_{t-1}$ is $\theta$-independent. Differentiating both sides w.r.t. $\theta$ gives
\[
    \frac{\partial y_{t-1}}{\partial \theta} = \frac{\partial \mu_\theta (y_t, t, x)}{\partial \theta} + \frac{\partial \mu_\theta(y_t, t, x)}{\partial y_t}\frac{\partial y_t}{\partial \theta}.
\]
Denote
\[
   A_t := \frac{\partial \mu_\theta (y_t, t, x)}{\partial \theta},\quad
   J_t := \frac{\partial \mu_\theta(y_t, t, x)}{\partial y_t},\quad
   G_t :=  \frac{\partial y_t}{\partial \theta}.
\]
Thus, we have
\[
    G_{t-1} = A_t + J_t G_t, \quad G_T = 0.
\]
Our final goal is:
\begin{align}
    \nabla_\theta R(\epsilon_{0:T}, \theta | x) &=\frac{\partial y_0}{\partial \theta}
    = G_0 \\
    &= A_1 + J_1 A_2 + J_1 J_2 A_3 + \cdots + J_1\cdots J_{T_1} A_t \\
    &= \sum_{s=1}^T \left(\prod_{u=1}^{s-1} J_u\right) A_s,
\end{align}
where we define $\prod_{u=1}^0 J_u := I$.
Then, we have
    \begin{align}
     \nabla_\theta{\E_{y \sim P_\theta(\cdot|x)} [f(y,z)]}
     &= \nabla_\theta \E_{\epsilon \sim P(\epsilon)}[f(R(\epsilon, \theta|x), z)] \\
     &= \E_{\epsilon \sim P(\epsilon)}[\nabla_\theta f(R(\epsilon, \theta|x), z)] \\
     &= \E_{\epsilon \sim P(\epsilon)}\left[ \nabla_\theta R(\epsilon, \theta \mid x)^\top \nabla_y f(R(\epsilon, \theta|x),z) \right] \\
     &= \E_{\epsilon \sim P(\epsilon)}\left[ \left(\sum_{s=1}^T \left(\prod_{u=1}^{s-1} J_u\right) A_s\right)^\top \nabla_y f(R(\epsilon, \theta|x),z) \right]
\end{align}
\end{proof}

\begin{lemma}[Gradient of Reparameterization method]
Assume the model prediction $y$ can be expressed as a transformation $y = T(\epsilon, \theta \mid x)$, $\epsilon \sim P(\epsilon)$. The total derivative of the decision objective $F$ w.r.t. $\theta$ can be computed as
\begin{align}
    \frac{d F}{d \theta} = - 
    \begin{bmatrix}
    \frac{d F}{d z^*} \\
    0 \\
    0
    \end{bmatrix}^\top 
    \begin{bmatrix}
    H & G^\top & Q^\top \\
    D(\lambda^*) G &  D(h(x, z^*)) & 0\\
    Q & 0 & 0
    \end{bmatrix}^{-1} 
    \begin{bmatrix}
    \E_{\epsilon \sim P(\epsilon)} [(\nabla_\theta T(\epsilon, \theta|x))^\top \nabla^2_{zy} f(z^*, y)]  \\
    0 \\
    0
    \end{bmatrix},
\end{align}
where $H = \E_{y \sim P_\theta(\cdot\mid x)} [\nabla_{zz}^2 f(y, z^*)] + \nabla_{zz}^2 (\lambda^{*\top} h(x, z^*))$ is the Hessian of the Lagrangian with respect to $z$, $G = \nabla_z h(x, z^*)$ is the Jacobian of the inequality constraints in $z^*$, and $Q = \nabla_z g(x, z^*)$ is the Jacobian of the equality constraints in $z^*$.
\end{lemma}

\begin{proof}
At the primal-dual optimal solution $(z_\theta^*, \lambda_\theta^*, \nu_\theta^*)$ to Eq.~\ref{eq:z_star_sto}, the following KKT conditions must hold:
\begin{align*}
     \nabla_z \calL(\theta, z_\theta^*, \lambda_\theta, \nu_\theta; x) &= 0,\\
     \lambda_\theta \odot h(x, z_\theta^*) &= 0, \\
     g(x, z_\theta^*) &= 0\\
     \lambda_\theta \geq 0, \nu_\theta &\geq 0, \\
     h(x, z_\theta^*) &\leq 0.
\end{align*}
Since $h$ does not depend on $\theta$ here, we can apply Proposition~\ref{prop:reparam} to the KKT conditions to get
\begin{align}
    &\frac{\partial \nabla_z \mathcal{L}}{\partial \theta} + \frac{\partial \nabla_z \mathcal{L}}{\partial z} \frac{\partial z^*}{\partial \theta} + \frac{\partial \nabla_z \mathcal{L}}{\partial \lambda^*} \frac{\partial \lambda^*}{\partial \theta} + \frac{\partial \nabla_z \mathcal{L}}{\partial \nu^*} \frac{\partial \nu^*}{\partial \theta} \notag \\
    &=  \E_{\epsilon \sim P(\epsilon)} [(\nabla_\theta T(\epsilon, \theta|x))^\top \nabla_y (\nabla_z f(z^*, y))] + \big(\E_{y \sim P_\theta(\cdot\mid x)} [\nabla_{zz}^2 f(z^*, y)] + \nabla_{zz}^2 h(x, z^*)\big) \frac{\partial z^*}{\partial \theta} \notag \\
    & \qquad + \nabla_{z} h(x, z^*) \frac{\partial \lambda^*}{\partial \theta} + \nabla_{z} g(x, z^*) \frac{\partial \lambda^*}{\partial \theta} \notag \\
    &= 0.
\end{align}
\begin{align}
    &\frac{\partial \lambda^* \odot h(x, z^*)}{\partial z^*} \frac{\partial z^*}{\partial \theta} + \frac{\partial \lambda^* \odot h(x, z^*)}{\partial \lambda^*} \frac{\partial \lambda^*}{\partial \theta} = D(\lambda^*) \nabla_z h(x, z^*) \frac{\partial z^*}{\partial \theta} + D(h(x, z^*))\frac{\partial \lambda^*}{\partial \theta} = 0.
\end{align}

In matrix form, we have
\begin{align}
    \begin{bmatrix}
    H & G^\top & Q^\top \\
    D(\lambda^*) G &  D(h(x, z^*)) & 0\\
    Q & 0 & 0
    \end{bmatrix} 
    \begin{bmatrix}
    \frac{\partial z^*}{\partial \theta}  \\
    \frac{\partial \lambda^*}{\partial \theta} \\
    \frac{\partial \nu^*}{\partial \theta}
    \end{bmatrix} = 
    - \begin{bmatrix}
    \E_{\epsilon \sim P(\epsilon)} [(\nabla_\theta T(\epsilon, \theta|x))^\top \nabla^2_{zy} f(z^*, y)]  \\
    0 \\
    0
    \end{bmatrix},
\end{align}
where $H = \E_{y \sim P_\theta(\cdot\mid x)} [\nabla_{zz}^2 f(z^*, y)] + \nabla_{zz}^2 (\lambda^{*\top} h(x, z^*)) + \nabla_{zz}^2 (\nu^{*\top}g(x, z^*))$, $G = \nabla_z h(x, z^*)$, and $Q = \nabla_z g(x, z^*)$. Furthermore, if equalities and inequalities are affine (as in main paper), $H$ reduces to $\E_{y \sim P_\theta(\cdot\mid x)} [\nabla_{zz}^2 f(y, z^*)]$ since $\nabla_{zz}^2h = \nabla_{zz}^2 g=0$.

By chain rule, we have
\begin{align*}
    \frac{d F}{d \theta} &= 
    \begin{bmatrix}
    \frac{d F}{d z^*} \\
    0
    \end{bmatrix}^\top 
    \begin{bmatrix}
    \frac{\partial z^*}{\partial \theta}  \\
    \frac{\partial \lambda^*}{\partial \theta}
    \end{bmatrix} \\ 
    &= - 
    \begin{bmatrix}
    \frac{d F}{d z^*} \\
    0 \\
    0
    \end{bmatrix}^\top 
    \begin{bmatrix}
    H & G^\top & Q^\top \\
    D(\lambda^*) G &  D(h(x, z^*)) & 0\\
    Q & 0 & 0
    \end{bmatrix}^{-1} 
    \begin{bmatrix}
    \E_{\epsilon \sim P(\epsilon)} [(\nabla_\theta T(\epsilon, \theta|x))^\top \nabla^2_{zy} f(z^*, y)]  \\
    0 \\
    0
    \end{bmatrix}.
\end{align*}
\end{proof}

\subsection{Proofs for Section~\ref{sec:score_fn}}\label{apd:proof_sf}
\begin{proposition}
Let $f:\calY \times \R^d \to \R$ be any function that does not depend on $\theta$. If $y \sim P_\theta(\cdot\mid x)$, then
\begin{align}
    \nabla_\theta{\E_{y \sim P_\theta(\cdot|x)} [f(y,z)]} = \E_{y \sim P_\theta(\cdot\mid x)} [f(y,z) {\frac{d\log P_\theta(y \mid x)}{d\theta}}].
\end{align}
\end{proposition}
\begin{proof}
\begin{align}
    \nabla_\theta{\E_{y \sim P_\theta(\cdot|x)} [f(y,z)]}
    &= \deriv{}{\theta} \E_{y \sim P_\theta(\cdot\mid x)} [f(y,z)] \\
    &= \deriv{}{\theta} \int f(y,z) P_\theta(y \mid x) \diff y \\
    &= \int P_\theta(y \mid x) \deriv{}{\theta} f(y,z) + f(y,z) \deriv{}{\theta} P_\theta(y \mid x) d y \\
    &= \int P_\theta(y \mid x) \deriv{}{\theta} f(y,z) + f(y,z) \deriv{}{\theta} \log P_\theta(y \mid x) *P_\theta(y \mid x) d y \label{eq:no_chain_rule} \\
    &= \E_{y \sim P_\theta(\cdot\mid x)} [\deriv{}{\theta} f(y,z)] + \E_{y \sim P_\theta(\cdot\mid x)} [f(y,z) {\frac{d\log P_\theta(y \mid x)}{d\theta}}].
\end{align}
This immediately implies the results by noticing $f$ does not depend on $\theta$.
\end{proof}

\begin{proposition}\label{prop:sf}
Let $P_\theta(y \mid x)$ be a probability density parameterized by $\theta \in \Theta$, and let $f: \calY \times \R^d \to \R$ be a scalar-valued function that does not depend on $\theta$. Fix any $z \in \R^d$. Suppose that there exists some neighborhood $N(\theta_0) \subseteq \Theta$ around $\theta_0 \in \Theta$ such that the following 3 assumptions are satisfied:
\begin{enumerate}[nosep]
    \item For all $\theta \in N(\theta_0)$, the function $h(y) := P_\theta(y \mid x)\, f(y,z)$ is integrable;
    \item For all $\theta \in N(\theta_0)$ and almost all $y \in \calY$, the gradient $\nabla_\theta P_\theta(y \mid x)$ exists; and
    \item There exists an integrable function $g: \calY \to \R$ that dominates $\nabla_\theta P_\theta(y \mid x)$. That is, for all $\theta \in N(\theta_0)$ and almost all $y \in \calY$, $\norm{\nabla_\theta P_\theta(y \mid x)}_1 \leq \abs{g(y)}$.
\end{enumerate}
Then,
\[
    \nabla_\theta{\E_{y \sim P_\theta(\cdot|x)} [f(y,z)]}
    = \E_{y \sim P_{\theta_0}(\cdot\mid x)} [f(y,z) \cdot \nabla_\theta \log P_{\theta_0}(y \mid x)].
\]
\end{proposition}
\begin{proof}
We make use of the log-derivative trick:
\[
    P_{\theta_0}(y \mid x) \cdot \nabla_\theta \log P_{\theta_0}(y \mid x)
    = \frac{P_{\theta_0}(y \mid x)}{P_{\theta_0}(y \mid x)} \cdot \nabla_\theta P_{\theta_0}(y \mid x)
    = \nabla_\theta P_{\theta_0} (y \mid x).
\]
Then
\begin{align*}
    \nabla_\theta \E_{y \sim P_{\theta_0}(\cdot\mid x)} [f(y,z)]
    &= \nabla_\theta \int_{\calY} f(y,z) P_{\theta_0}(y \mid x) \diff y \\
    &= \int_{\calY} \nabla_\theta \left[ f(y,z)\, P_{\theta_0}(y \mid x) \right] \diff y
        & \text{Leibniz integral rule} \\
    &= \int_{\calY} f(y,z) \, P_{\theta_0}(y \mid x) \, \nabla_\theta \log P_{\theta_0}(y \mid x) \diff y
        & \text{log-derivative trick} \\
    &= \E_{y \sim P_{\theta_0}(\cdot\mid x)} \left[ f(y,z) \nabla_\theta \log P_{\theta_0}(y \mid x) \right].
\end{align*}
\end{proof}

\begin{lemma}[Gradient of Score Function]\label{lm:sf}
The total derivative of the decision objective $F$ w.r.t. $\theta$ can be computed as
\begin{align}
    \frac{d F}{d \theta} &= - 
    \begin{bmatrix}
    \frac{d F}{d z^*} \\
    0 \\
    0
    \end{bmatrix}^\top 
    \begin{bmatrix}
    H & G^\top & Q^\top \\
    D(\lambda^*) G &  D(h(x, z^*)) & 0\\
    Q & 0 & 0
    \end{bmatrix}^{-1} 
    \begin{bmatrix}
    \E_{y \sim P_\theta(\cdot\mid x)} [\nabla_z f(z^*, y) (\frac{dELBO}{d\theta})^\top]  \\
    0 \\
    0
    \end{bmatrix}.
\end{align}
\end{lemma}
\begin{proof}
Differentiate this KKT system w.r.t. $\theta$ and applying Proposition~\ref{prop:sf} yields
\begin{align}
    &\frac{\partial \nabla_z \mathcal{L}}{\partial \theta} + \frac{\partial \nabla_z \mathcal{L}}{\partial z} \frac{\partial z^*}{\partial \theta} + \frac{\partial \nabla_z \mathcal{L}}{\partial \lambda^*} \frac{\partial \lambda^*}{\partial \theta} + \frac{\partial \nabla_z \mathcal{L}}{\partial \nu^*} \frac{\partial \nu^*}{\partial \theta} \notag \\
    &= \E_{y \sim P_\theta(\cdot\mid x)} [\nabla_z f(z^*, y)(\nabla_\theta \log P_\theta (y|x))^\top] + (\E_{y \sim P_\theta(\cdot\mid x)} [\nabla_{zz}^2 f(z^*, y)]  + \nabla_{zz}^2 (\lambda^{\star\top} h(x, z^*)) )\frac{\partial z^*}{\partial \theta} \notag \\
    &\quad + \nabla_{z} h(x, z^*) \frac{\partial \lambda^*}{\partial \theta} + \nabla_z g(x, z^*)\frac{\partial \nu^*}{\partial \theta} \notag  \\
    &= 0.
\end{align}
\begin{align}
    &\frac{\partial \lambda^* \odot h(x, z^*)}{\partial z^*} \frac{\partial z^*}{\partial \theta} + \frac{\partial \lambda^* \odot h(x, z^*)}{\partial \lambda^*} \frac{\partial \lambda^*}{\partial \theta} \notag \\
    =& D(\lambda^*) \nabla_z h(x, z^*) \frac{\partial z^*}{\partial \theta} + D(h(x, z^*))\frac{\partial \lambda^*}{\partial \theta} = 0.
\end{align}

In matrix form, this becomes
\begin{equation}
    \begin{bmatrix}
    H & G^\top & Q^\top \\
    D(\lambda^*) G &  D(h(x, z^*)) & 0\\
    Q & 0 & 0
    \end{bmatrix} 
    \begin{bmatrix}
    \frac{\partial z^*}{\partial \theta}  \\
    \frac{\partial \lambda^*}{\partial \theta} \\
    \frac{\partial \nu^*}{\partial \theta}
    \end{bmatrix} = 
    - \begin{bmatrix}
    \E_{y} [\nabla_z f(z^*, y) (\nabla_\theta \log P_\theta(y \mid x))^\top]  \\
    0 \\
    0
    \end{bmatrix},
\end{equation}
where $H = \E_{y \sim P_\theta(\cdot\mid x)} [\nabla_{zz}^2 f(z^*, y)] + \nabla_{zz}^2 (\lambda^{*\top} h(x, z^*))$, $G = \nabla_z h(x, z^*)$.

Applying the chain rule to $F$ now gives
\begin{align}
    \frac{d F}{d \theta} &= 
    \begin{bmatrix}
    \frac{d F}{d z^*} \\
    0 \\
    0
    \end{bmatrix}^\top 
    \begin{bmatrix}
    \frac{\partial z^*}{\partial \theta}  \\
    \frac{\partial \lambda^*}{\partial \theta} \\
    \frac{\partial \nu^*}{\partial \theta}
    \end{bmatrix} \\ 
    &= -
    \begin{bmatrix}
    \frac{d F}{d z^*} \\
    0 \\
    0
    \end{bmatrix}^\top 
    \begin{bmatrix}
    H & G^\top & Q^\top \\
    D(\lambda^*) G &  D(h(x, z^*)) & 0\\
    Q & 0 & 0
    \end{bmatrix}^{-1}
    \begin{bmatrix}
    \E_{y \sim P_\theta(\cdot\mid x)} [\nabla_z f(z^*, y) (\nabla_\theta \log P_\theta(y \mid x))^\top]  \\
    0 \\ 
    0
    \end{bmatrix}.
\end{align}

Then, we replace $\nabla_\theta \log P_\theta(y \mid x)$ with the gradient of ELBO score for sample $y$ and have
\begin{align}
    \frac{d F}{d \theta} &= - 
    \begin{bmatrix}
    \frac{d F}{d z^*} \\
    0 \\
    0
    \end{bmatrix}^\top 
    \begin{bmatrix}
    H & G^\top & Q^\top \\
    D(\lambda^*) G &  D(h(x, z^*)) & 0\\
    Q & 0 & 0
    \end{bmatrix}^{-1} 
    \begin{bmatrix}
    \E_{y \sim P_\theta(\cdot\mid x)} [\nabla_z f(z^*, y) (\frac{dELBO}{d\theta})^\top]  \\
    0 \\
    0
    \end{bmatrix}.
\end{align}
\end{proof}

\begin{remark}[Why we cannot compute the gradient using score-matching]
    One may attempt to apply the chain rule $\nabla_\theta \log P_\theta(y|x) =\nabla_\theta y \nabla_y \log P_\theta(y|x)$, and then estimate $\nabla_y \log P_\theta(y_{t+1}|x) \approx \nabla_y \log P_\theta(y_{t+1}|y_t, x) \approx s_\theta(y_t, t, x)$ via score-matching~\citep{song_score-based_2020} (using the learned score $s_\theta(y_t, t, x)$ of the diffusion model). However, this approach is invalid in our setting: Under the log-trick, $y$ is treated as a free variable and $\theta$ enters only through $ P_\theta (y|x)$, so the pathwise term $\nabla_{\theta}y$ does not exist (see~\Cref{apd:proof_sf} for derivation). Our ELBO-based surrogate (Eq.~\eqref{eq:elbo_grad_approx}) avoids this obstacle entirely.
\end{remark}

\subsection{Proof for Eq.~\ref{eq:weighted_elbo}}\label{apd:prf_IS_elbo}
Based on the results in \Cref{lm:sf}, we have
\begin{align}
    \frac{d F}{d \theta} 
    &= - 
    \begin{bmatrix}
    \frac{d F}{d z^*} \\
    0 \\
    0
    \end{bmatrix}^\top 
    \begin{bmatrix}
    H & G^\top & Q^\top \\
    D(\lambda^*) G &  D(h(x, z^*)) & 0\\
    Q & 0 & 0
    \end{bmatrix}^{-1} \begin{bmatrix}
    \E_{y \sim P_\theta(\cdot|x)} [\nabla_z f(z^*, y) (\frac{d\ELBO}{d\theta})^\top]  \\
    0 \\
    0
    \end{bmatrix} \\
    &= \underbrace{- 
    \begin{bmatrix}
    \frac{d F}{d z^*} \\
    0 \\
    0
    \end{bmatrix}^\top 
    \begin{bmatrix}
    H & G^\top & Q^\top \\
    D(\lambda^*) G &  D(h(x, z^*)) & 0\\
    Q & 0 & 0
    \end{bmatrix}^{-1}}_{:= u(\theta)^\top} \frac{d}{d \theta}\begin{bmatrix}
    \E_{y \sim P_\theta(\cdot|x)} [\nabla_z f(z^*, y) \ELBO]  \\
    0 \\
    0
    \end{bmatrix}\\
    &=\frac{d}{d \theta} \E_{y \sim P_\theta(\cdot|x)}[ \underbrace{u(\theta)^\top \begin{bmatrix}
    [\nabla_z f(z^*, y)]  \\
    0 \\
    0
    \end{bmatrix}}_{:=w_\theta(y)} \ELBO] \\
    &= \frac{d}{d \theta} \E_{y \sim P_\theta(\cdot|x)}[ {\mathrm{detach}[w_\theta(y)]}{\ELBO}]
\end{align}

\subsection{Empirical evidence for ELBO gradient approximation}
Assume our model is $\theta = (A, B, c)$, and the noise is predicted by $\epsilon_\theta(y_t, t, x) = A_t y_t + B_t x + c_t$. True data $y \sim \calN(Wx, I)$. In this way, there is a closed-form solution for true $\nabla_\theta \log p(y_0|x)$ since $y_0$ is a Gaussian and $y_t = \sqrt{\bar \alpha_t}y_0 + \sqrt{1 - \bar \alpha_t}\epsilon$.

\begin{figure}[htbp]
    \centering
    \includegraphics[width=0.6\linewidth]{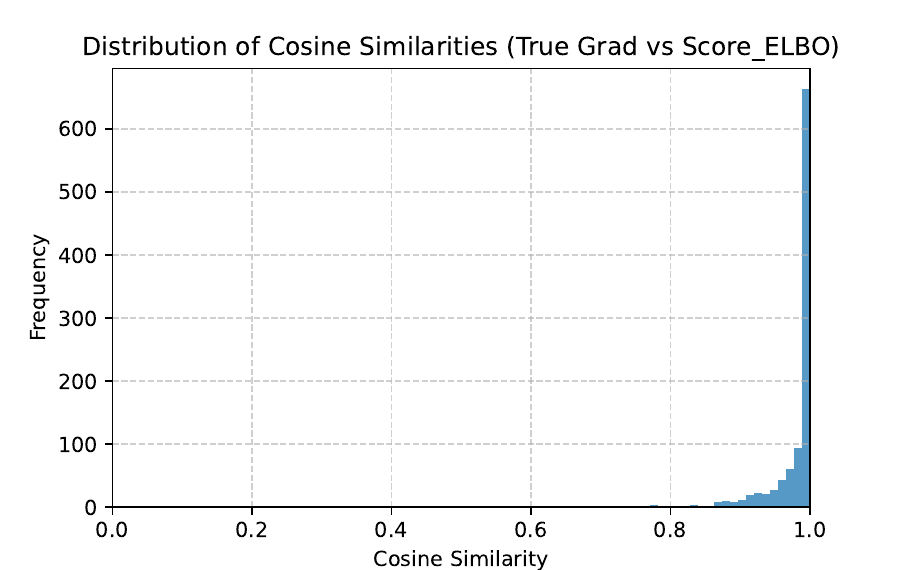}
    \caption{Compare the cosine similarity between the true gradient and the estimated gradient using a linear model.}
    \label{fig:placeholder}
\end{figure}

\subsection{Deterministic Optimization and Gaussian Model in Stochastic Optimization}
\paragraph{Deterministic Optimization}\label{apd:det}
Since deterministic can also be viewed as a reparameterization trick without any randomness $\epsilon$, we can reuse our derivation in Section~\ref{sec:reparm} and compute the gradient $\nabla_x F(x)$ by Eq.~\ref{eq:total_grad_reparam}.

\paragraph{Gaussian Model in Stochastic Optimization}\label{apd:gs}
Since there are many recent papers that use the Gaussian model as a predictor for stochastic DFL, we also claim that it has the reparameterization and score function form.

\begin{enumerate}
    \item Reparameterization with Gaussian Model. Also using reparameterization trick:
    \begin{align}
        \nabla_\theta{\E_{y \sim P_\theta(\cdot|x)} [f(y,z)]} = \E_{\epsilon \sim P(\epsilon)} [(\nabla_\theta y \nabla_y f(y, z)].
    \end{align}
    But with the predictor instantiated as a Gaussian model, i.e., $y$ is computed by $y = R(\epsilon, \theta|x))^\top$ and $R$ is a Gaussian model.

    \item Score Function with Gaussian Model. Recall that we need to approximate the term $\log P_\theta (y|x)$ for the diffusion model. However, this term has a closed-form for a Gaussian model. Assume our Gaussian model is $\calN (\mu_\theta, \Sigma_\theta)$, then the negative log-likelihood is
\begin{align}
    \log P_\theta(y \mid x) = - \frac{1}{2} [ (y - \mu_\theta)^\top \Sigma_\theta^{-1} (y - \mu_\theta) + \log \det \Sigma_\theta + d \log (2\pi)].
\end{align}
Then, the gradient for the score function can be calculated using $\nabla_\theta \log P_\theta(y \mid x)$.
Gaussian models are powerful tools for many DFL tasks. However, we want to claim that the diffusion model is more general and requires less model tuning and model assumptions.

\end{enumerate}

\subsection{Training details}
We summarize our model settings for the deterministic model, Gaussian model, and diffusion model in~\Cref{tab:model_arch}.

\begin{table}[htbp]
\centering
\small
\setlength{\tabcolsep}{6pt}
\renewcommand{\arraystretch}{1.2}
\begin{tabularx}{\linewidth}{lXXX}
\toprule
\textbf{Parameter} & \textbf{Deterministic MLP} & \textbf{Gaussian MLP} & \textbf{Diffusion} \\
\midrule
Trunk (layers $\times$ width) &
$2 \times 1024$ &
$2 \times 1024$ &
$2 \times 1024$ \\
Activation &
ReLU / Swish (SiLU) &
ReLU / Swish (SiLU) &
Swish (SiLU) \\
Inputs to network &
$\;x \in \mathbb{R}^{d_x}$ &
$\;x \in \mathbb{R}^{d_x}$ &
$\;[\,y_t \in \mathbb{R}^{d_y},\; t,\; x\,]$ \\
Time embedding &
--- &
--- &
Sinusoidal $t$ (16-d) $\rightarrow$ 2 FC + SiLU \\
Output head &
$\hat y \in \mathbb{R}^{d_y}$ &
$\mu_\theta(x), \log \sigma_\theta^2(x) \in~\mathbb{R}^{d_y}$ &
$\hat\epsilon=\epsilon_\theta(y_t,t,x) \in \mathbb{R}^{d_y}$ \\
Uncertainty form &
None (point) &
Gaussian $\mathcal N(\mu,\mathrm{diag}(\sigma^2))$ &
Non-parametric (learned $P_\theta(y\mid x)$) \\
Two-stage training loss &
MSE on $y$ &
Gaussian NLL &
Weighted denoising MSE (ELBO-equiv.) \\
DFL gradient computation &
Implicit differentiation through KKT&
(1) Reparam: $y=\mu+\sigma\odot\epsilon$ + implicit diff; (2) Gaussian score-function &
(1) Pathwise (backprop through sampler, $T$ steps) + implicit diff; (2) \emph{Weighted-ELBO} score-function\\
Sample size $M$&
--- &
$10$ (synthetic), $25$ (power), $50$ (portfolio) &
$10$ (synthetic), $25$ (power), $50$ (portfolio) \\
Learning rate (lr) &
Task-specific, mostly $1 \times10^{-4}$ &
Task-specific, mostly $1 \times10^{-4}$ &
reparam: $1\times10^{-5}$;\; score-fn: $8\times10^{-6}$\\
Inference &
Use one $\hat y$ in optimizer &
Draw $M$ samples $y^{(m)}$ from Gaussian model &
Reverse diffusion to sample $M$ samples $y^{(m)}$  \\
\bottomrule
\end{tabularx}
\vspace{2mm}
\caption{Comparison of deterministic, Gaussian, and diffusion model architectures.}
\label{tab:model_arch}
\end{table}

For all experiments, we perform 10 random seeds to evaluate variability. We also find that adding a small regularizer during DFL training can help the model learning the data distribution and avoid some bad local minima, leading to a stable training process.

\subsection{Details of synthetic example}
In this example, we consider a factory that decides how much to manufacture for each of $d \in \N$ products. The parameter $Y \in \R^d$ represents the \textit{profit margin} for each product, i.e., $Y_i$ is the profit per unit of product $i$; due to uncertainty in market conditions, $Y$ is uncertain. The factory's decision $z \in [0,C]^d$ represents how much of each product to manufacture, where $C$ is the maximum capacity for each product. For simplicity, we do not consider any contextual features $x$ in this example. That means DFL learns a distribution that generates $y$ that can minimize the decision objective.

Suppose that the factory has a risk-averse cost function $f(y,z) = \exp(-y^\top z)$, which indicates that the factory wants to put a larger weight on the product with higher profit $Y_i$. Intuitively, if the factory knew $Y$ exactly, then the optimal strategy would be all-or-nothing: set $z_i = C$ if $Y_i > 0$, or $z_i = 0$ if $Y_i < 0$. Likewise, with respect to a point prediction of $Y$, the optimal deterministic decision $z_\mathrm{det}^* \in \{0, C\}^d$ is attained on the boundary of the feasible set.

Under uncertainty, the decision-maker seeks to minimize the \textbf{expected cost} by solving a stochastic optimization problem:
\begin{align}
    z_\mathrm{sto}^* &\in \argmin_{z \in [0,C]^d} \E_{y \sim P_\theta(\cdot|x)} [\exp(-y^\top z)].
\end{align}
In this stochastic case, the optimal investment $z_\mathrm{sto}^*$ typically lies in the interior of the feasible region, which balances the potential high reward of investing against the risk of losses.

Then, we compute the necessities for diffusion DFL:
\begin{align}
    H &= \E_{y \sim P_\theta(\cdot\mid x)} [\nabla_{zz}^2 g(z^*, y)] + (\lambda^*)^\top \nabla_{zz}^2 h(x, z^*) = \exp(-y^\top z) y y^\top \\
    G &= \nabla_z h(x, z^*) = -\exp(-y^\top z) y.
\end{align}
For reparameterization, we have
\begin{align*}
    (\frac{d loss}{d \theta})^\top &= 
    \begin{bmatrix}
    \frac{d loss}{d z^*} \\
    0
    \end{bmatrix}^\top 
    \begin{bmatrix}
    \frac{\partial z^*}{\partial \theta}  \\
    \frac{\partial \lambda^*}{\partial \theta}
    \end{bmatrix} \\ 
    &= - 
    \begin{bmatrix}
    \frac{d loss}{d z^*} \\
    0
    \end{bmatrix}^\top 
    \begin{bmatrix}
    H & G^\top \\
    D(\lambda^*) G &  D(h(x, z^*))
    \end{bmatrix} ^{-1} \begin{bmatrix}
    \frac{1}{M} \sum_{i=1}^M (\nabla_\theta y_i)^\top \nabla_{zy}^2 g(z^*, y_i)  \\
    0
    \end{bmatrix}
\end{align*}
where $\nabla_{zy}^2 g(z^*, y) = \exp(-y^\top z) (yz^\top - I_d)$ in this case.

For the score function, we have 
\begin{align*}
    (\frac{d loss}{d \theta})^\top
    &= - 
    \begin{bmatrix}
    \frac{d loss}{d z^*} \\
    0
    \end{bmatrix}^\top 
    \begin{bmatrix}
    H & G^\top \\
    D(\lambda^*) G &  D(h(x, z^*))
    \end{bmatrix} ^{-1} \begin{bmatrix}
    \E_{y} [\nabla_z g(z^*, y) (\frac{dELBO}{d\theta})^\top]  \\
    0
    \end{bmatrix} \\
    &\approx - \begin{bmatrix}
    \frac{d loss}{d z^*} \\
    0
    \end{bmatrix}^\top 
    \begin{bmatrix}
    H & G^\top \\
    D(\lambda^*) G &  D(h(x, z^*))
    \end{bmatrix} ^{-1} \begin{bmatrix}
    \frac{1}{M} \sum_{i=1}^M \nabla_z g(z^*, y_i) (\frac{dELBO_i}{d\theta})^\top  \\
    0
    \end{bmatrix}.
\end{align*}

\subsection{Details on power schedule task}
This task involves a 24-hour electricity generation scheduling problem with uncertain demand. The decision $z \in \R^{24}$ represents the electricity output to schedule for each hour of the next day. The uncertainty $y \in \R^{24}$ represents the actual power demand for each of the 24 hours. The goal is to meet demand as closely as possible at minimum cost. We also consider a decision cost function that penalizes storage, excess generation, and ramping following~\cite{donti_task-based_2017}:
\begin{enumerate}
    \item Let $\gamma_s$ and $\gamma_a$ be the per-unit costs of shortage (not meeting demand) and excess (over-generation), respectively. We use $\gamma_s = 50$ and $\gamma_e = 0.5$ in our experiment.
    \item Let $c_{r}$ be a penalty on hour-to-hour changes in generation. We use $c_r = 0.4$ in appropriate units.
\end{enumerate}
Formally, if $z = (z_1,\dots, z_{24})$ and $y = (y_1, \dots, y_{24})$, the loss for a single day is
\begin{align}
    &\min_z \  \E_{y \sim P_\theta(\cdot|x)}[f(y, z)] = \sum_{i=1}^{24} \E_{y \sim P_\theta(\cdot|x)} [\gamma_s[y_i - z_i]_+ + \gamma_e[z_i - y_i]_+ + \frac{1}{2} (z_i - y_i)^2] \notag \\
    &\ \mathrm{s.t.} \ |z_i - z_{i - 1}| \leq c_r \ \text{for all} \ i \in \{1, 2, \dots, 24\}.
\end{align}
Then, we compute the necessities for diffusion DFL:
\begin{align}
    H &= \E_{y \sim P_\theta(\cdot\mid x)} [\nabla_{zz}^2 g(z^*, y)] + (\lambda^*)^\top \nabla_{zz}^2 h(x, z^*) = I_n \\
    G &= \nabla_z h(x, z^*) = \begin{bmatrix}
        -1 &  1 &  0 & \cdots & 0 \\
         0 & -1 &  1 & \cdots & 0 \\
        \vdots & & \ddots & \ddots & \vdots \\
         0 & \cdots & 0 & -1 & 1
        \end{bmatrix}.
\end{align}
For reparameterization, we have
\begin{align*}
    (\frac{d loss}{d \theta})^\top &= 
    \begin{bmatrix}
    \frac{d loss}{d z^*} \\
    0
    \end{bmatrix}^\top 
    \begin{bmatrix}
    \frac{\partial z^*}{\partial \theta}  \\
    \frac{\partial \lambda^*}{\partial \theta}
    \end{bmatrix} \\ 
    &= - 
    \begin{bmatrix}
    \frac{d loss}{d z^*} \\
    0
    \end{bmatrix}^\top 
    \begin{bmatrix}
    H & G^\top \\
    D(\lambda^*) G &  D(h(x, z^*))
    \end{bmatrix} ^{-1} \begin{bmatrix}
    \frac{1}{M} \sum_{i=1}^M (\nabla_\theta y_i)^\top \nabla_{zy}^2 g(z^*, y_i)  \\
    0
    \end{bmatrix}
\end{align*}
where $\nabla_{zy}^2 g(z^*, y) = -I$ in this case.

For the score function, we have 
\begin{align*}
    (\frac{d loss}{d \theta})^\top
    &= - 
    \begin{bmatrix}
    \frac{d loss}{d z^*} \\
    0
    \end{bmatrix}^\top 
    \begin{bmatrix}
    H & G^\top \\
    D(\lambda^*) G &  D(h(x, z^*))
    \end{bmatrix} ^{-1} \begin{bmatrix}
    \E_{y} [\nabla_z g(z^*, y) (\frac{dELBO}{d\theta})^\top]  \\
    0
    \end{bmatrix} \\
    &\approx - \begin{bmatrix}
    \frac{d loss}{d z^*} \\
    0
    \end{bmatrix}^\top 
    \begin{bmatrix}
    H & G^\top \\
    D(\lambda^*) G &  D(h(x, z^*))
    \end{bmatrix} ^{-1} \begin{bmatrix}
    \frac{1}{M} \sum_{i=1}^M \nabla_z g(z^*, y_i) (\frac{dELBO_i}{d\theta})^\top  \\
    0
    \end{bmatrix}.
\end{align*}

\paragraph{Dataset.} We use real historical electricity load data from the PJM regional grid (a standard public dataset)~\cite{pjm_interconnection_data_2025}. The features $x$ for each day include: the previous day's 24-hour load profile, the previous day's temperature profile, calendar features, and seasonal sinusoidal features. In total, $d_x = 28$ features for each day were constructed. We normalized all input features for training. The target label $y$ is the next day's 24-hour load vector.

\begin{figure}[htbp]
    \centering
    \includegraphics[width=0.98\linewidth]{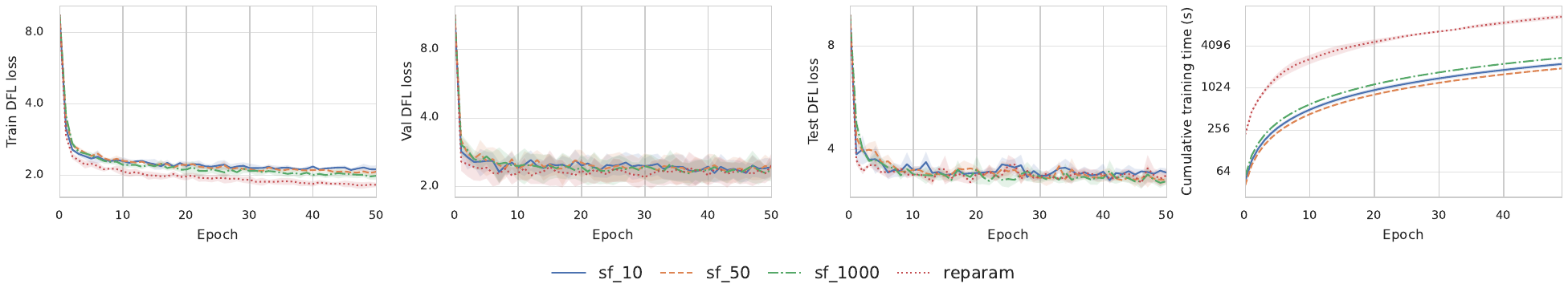}
    \caption{Comparison between different sample sizes for score function and reparameterization.}
    \label{fig:abl_resample_full}
\end{figure}

For completeness, we include an extended comparison of different sample sizes in \Cref{fig:abl_resample_full}, which further highlights that additional samples yield diminishing returns in accuracy while linearly increasing compute cost.

\begin{figure}[htbp]
    \centering
    \includegraphics[width=0.98\linewidth]{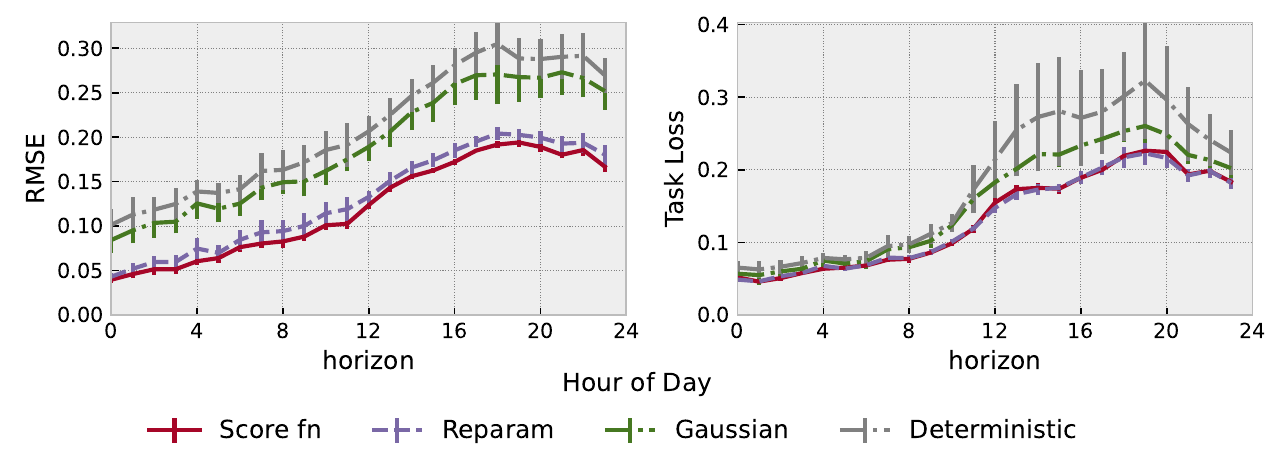}
    \caption{Results on the 24-hour power grid scheduling task.}
    \label{fig:power_grid_24_hour}
\end{figure}

\begin{figure}[htbp]
    \centering
    \includegraphics[width=0.98\linewidth]{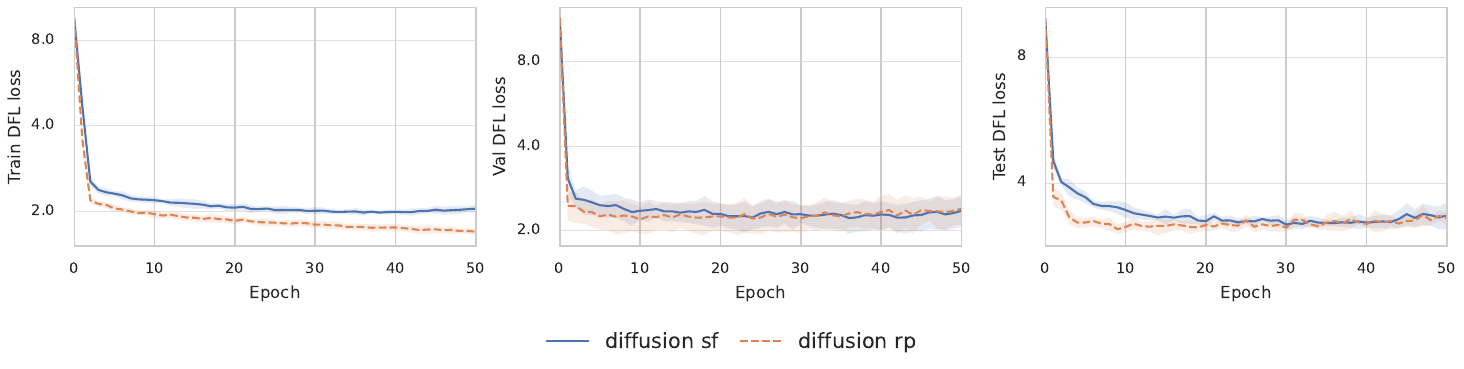}
    \caption{Score function vs. Reparameterization}
    \label{fig:power_sf_vs_rp}
\end{figure}

\subsection{Details on stock portfolio task}\label{apd:stock}
We consider a mean-variance portfolio optimization problem with uncertain returns. Here $y \in \R^n$ represents the random next-day returns of $n$ assets (stocks), and $z \in \R^n$ are the portfolio weights we assign to each asset (the fraction of our capital invested in each stock).
Our goal is to maximize expected return while keeping the risk (variance) low. This can be written as minimizing a loss that is a negative expected return plus a quadratic penalty on variance:
\begin{align}
    \min_z \ \E_{y \sim P_\theta(\cdot|x)}[f(y, z)] = \E_{y \sim P_\theta(\cdot|x)} \left[\frac{\alpha}{2} z^\top y y^\top z - y^\top z\right], \quad \mathrm{s.t.}\quad z^\top \mathbf{1} = 1, \ 0 \leq z_i \leq 1.
\end{align}
Then, we compute the necessities for diffusion DFL:
\begin{align}
    &H = \E_{y \sim P_\theta(\cdot\mid x)} [\nabla_{zz}^2 f(z^*, y)] + (\lambda^*)^\top \nabla_{zz}^2 h(x, z^*) + \nabla_{zz}^2 (\nu^{*\top}g(x, z^*)) = \alpha \E_{y \sim P_\theta(y \mid x)} [y y^\top], \\
    &G = \nabla_z h(x, z^*) = \begin{bmatrix}
        I_n, \\
        -I_n
    \end{bmatrix}\\
    &Q = \nabla_z g(x, z^*) = \mathbf{1}^\top.
\end{align}
For reparameterization, we have
\begin{align*}
    (\frac{d loss}{d \theta})^\top &= 
    \begin{bmatrix}
    \frac{d loss}{d z^*} \\
    0\\
    0
    \end{bmatrix}^\top 
    \begin{bmatrix}
    \frac{\partial z^*}{\partial \theta}  \\
    \frac{\partial \lambda^*}{\partial \theta} \\
    \frac{\partial \nu^*}{\partial \theta}
    \end{bmatrix} \\ 
    &\approx - 
    \begin{bmatrix}
    \frac{d loss}{d z^*} \\
    0\\
    0
    \end{bmatrix}^\top 
    \begin{bmatrix}
    H & G^\top & Q^\top \\
    D(\lambda^*) G &  D(h(x, z^*)) & 0\\
    Q & 0 & 0
    \end{bmatrix}^{-1} \begin{bmatrix}
    \frac{1}{M} \sum_{i=1}^M (\nabla_\theta y_i)^\top \nabla_{zy}^2 g(z^*, y_i)  \\
    0 \\
    0
    \end{bmatrix}
\end{align*}
where $\nabla_{zy}^2 f(z^*, y) = \E_{y} [2\alpha y^\top z - 1]$ in this case. 

For the score function, we have 
\begin{align*}
    (\frac{d loss}{d \theta})^\top
    &= \begin{bmatrix}
    \frac{d loss}{d z^*} \\
    0\\
    0
    \end{bmatrix}^\top 
    \begin{bmatrix}
    H & G^\top & Q^\top \\
    D(\lambda^*) G &  D(h(x, z^*)) & 0\\
    Q & 0 & 0
    \end{bmatrix}^{-1} \begin{bmatrix}
    \E_{y} [\nabla_z g(z^*, y) (\frac{dELBO}{d\theta})^\top]  \\
    0\\
    0
    \end{bmatrix} \\
    &\approx - \begin{bmatrix}
    \frac{d loss}{d z^*} \\
    0\\
    0
    \end{bmatrix}^\top 
    \begin{bmatrix}
    H & G^\top & Q^\top \\
    D(\lambda^*) G &  D(h(x, z^*)) & 0\\
    Q & 0 & 0
    \end{bmatrix}^{-1}  \begin{bmatrix}
    \frac{1}{M} \sum_{i=1}^M \nabla_z g(z^*, y_i) (\frac{dELBO_i}{d\theta})^\top  \\
    0\\
    0
    \end{bmatrix},
\end{align*}
where $\nabla_z g(z^*, y_i) = \alpha y_i y_i^\top z^* - y_i$.

\paragraph{Dataset.} We use daily stock prices from 2004–2017 for constituents of the S\&P 500 index~\citep{quandl_wiki_dataset_nasdaq_2025}. We obtained this data via Quandl’s API (specifically WIKI pricing data; the user will need a Quandl API key to replicate. We compute daily returns for each stock (percentage change). To construct features $x$, we use a rolling window of recent history for each asset. Specifically, for each day, a data point is for predicting next day’s returns: we include the past 5 days of returns for each of the $n$ assets, past 5 days of trading volume for each asset, plus some aggregate features. To avoid an explosion of dimension with large $n$, we also include PCA-compressed features: we take the top principal components of the last 5-day return matrix to summarize cross-asset trends. In the end, for $n=50$ assets, we ended up with $d_x = 28$ features. All features are normalized and we use a time-series split: first 70\% of days for training (2004–2013), next 10\% for validation (2014), last 20\% for test (2015–2017). We evaluate performance on the test set by simulating the portfolio selection every day and computing the average return achieved.

\subsection{(Additional task) Details on inventory stock problem}\label{apd:toy}
We also validate our approaches on a toy inventory control problem. In this task, the uncertain demand $y$ is drawn from a multi-modal distribution (a mixture of Gaussians), where we vary the number of mixture components $K$ to control the distribution complexity:
\begin{align}
    p(x) = \sum_{j=1}^K \pi_j \phi(x;\mu_j, \Sigma_j),
\end{align}
where $\pi_j$ is the probability of choosing component j and $\phi(x;\mu_j, \Sigma_j)$ is a multivariate Gaussian density with parameter $(\mu_j, \Sigma_j)$.
The cost function follows the standard newsvendor formulation with piecewise penalties for under-stock and over-stock:
\begin{align}\label{eq:obj_inventory}
    f(y, z) = c_0z + \frac{1}{2} q_0 z^2 + c_b [y - z]_{+} + \frac{1}{3} r_b ([y - z]_{+}^3) + c_h [z - y]_{+} + \frac{1}{3} r_h ([z - y]_{+}^3).
\end{align}
Our learning objective is to minimize the expected cost over this stochastic demand, i.e., a stochastic optimization problem:
\begin{align}
    \min_z L(\theta) = \E_{y \sim P(\cdot|x)} [f(y, z)] \quad \mathrm{s.t.} \ 0 \leq z \leq z_{max}.
\end{align}

\begin{figure}[htbp]
    \centering
    \includegraphics[width=0.98\linewidth]{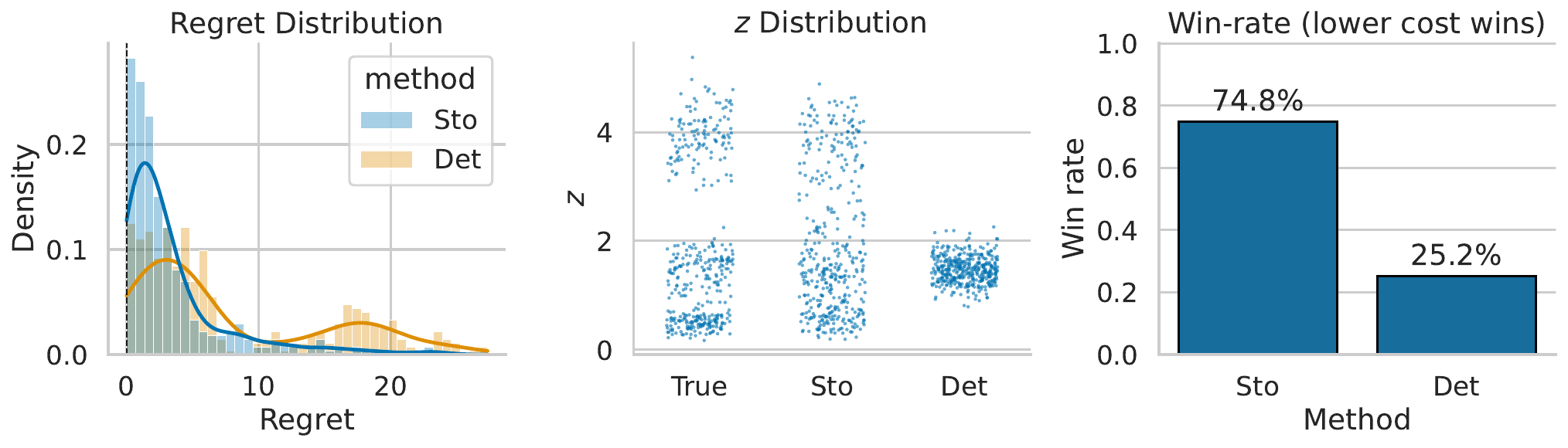}
    \caption{Toy decision task comparing deterministic and diffusion DFL. \emph{Left}: distribution of per-instance regret (lower is better). \emph{Middle}: distribution of chosen decision $z$ in the lower-level; the stochastic method tracks the true distribution $z^*$ more closely. \emph{Right}: pairwise win-rate on test set; a large fraction of costs from the stochastic method are lower than the deterministic one, indicating that modeling uncertainty yields better decisions.}
    \label{fig:toy_regret}
\end{figure}

% \begin{figure}[htbp]
%     \centering
%     \includegraphics[width=0.98\linewidth]{exp_res/power_grid/pretrain_after_pretrain.pdf}
%     \caption{Diffusion DFL vs. diffusion two-stage.}
%     \label{fig:toy_two_stage}
% \end{figure}

\begin{figure}[htbp]
    \centering
    \includegraphics[width=0.5\linewidth]{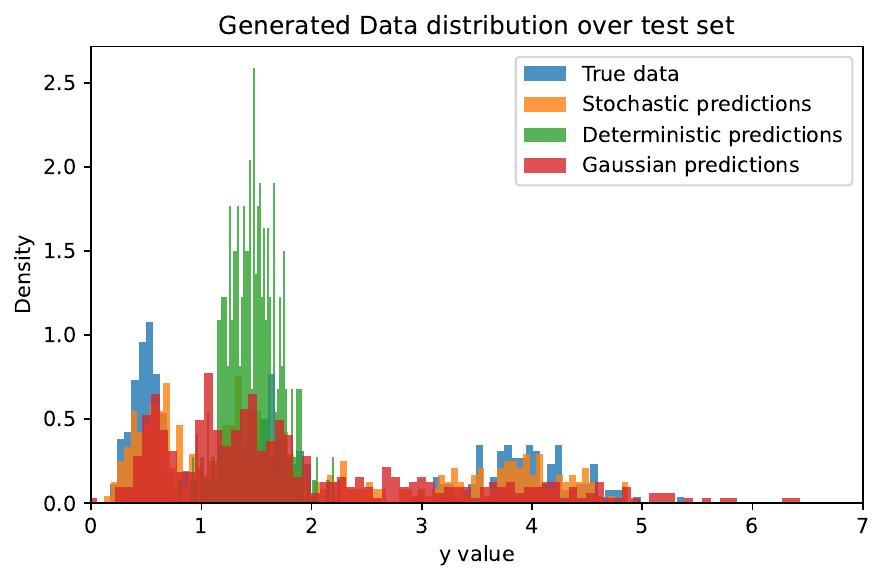}
    \caption{Prediction distribution for inventory stock problem.}
    \label{fig:toy_gen_distr}
\end{figure}

We compare a deterministic DFL model against our diffusion DFL model on this toy task. \Cref{fig:toy_regret} summarizes the results, where the diffusion model (stochastic DFL) achieves substantially lower regret on average than the deterministic model. Besides, we observe that the decision $z$ obtained by our diffusion method closely tracks the true optimal decisions $z^*$ by capturing the multi-modal demand uncertainty, whereas the deterministic predictor's decisions deviate more. In \Cref{fig:toy_regret} (c), we directly compare the decision outcomes via a win-rate: the fraction of test instances where one method achieves lower cost than the other. The diffusion DFL method attains a win-rate of about 75\% against the deterministic baseline, which confirms that modeling uncertainty leads to better downstream decisions

For data generation, we set $\mu = [-4, 0, 4], \Sigma = [0.15, 0.25, 0.15]^\top \mathbf{1}$ for $K=3$, $\mu = [-6., -3., 0., 3., 6.], \Sigma = [0.15, 0.25, 0.35, 0.25, 0.15]^\top \mathbf{1}$ for $K=5$, and $\mu = [-8.0, -6.0, -4.0, -2.0, -1.0, 0.0, 1.2,  2.8,  4.5,  7.5], \Sigma = [0.30, 0.75, 0.25, 0.40, 0.22, 0.20, 0.22, 0.35, 0.70, 1.25]^\top \mathbf{1}$.

Following our previous derivation, we can compute the necessities for diffusion DFL by
\begin{align}
    &H = \E_{y \sim P_\theta(\cdot\mid x)} [\nabla_{zz}^2 f(z^*, y)] + (\lambda^*)^\top \nabla_{zz}^2 h(x, z^*) = diag(q_0) + q_b I_{(y>z)} + q_h I_{z>y} \\
    &G = \nabla_z h(x, z^*) = \begin{bmatrix}
        -I\\
        I
    \end{bmatrix}, \\
    &D(h(x, z^*)) = 0.
\end{align}

\subsection{Additional Related Works}
\paragraph{Stochastic optimization}
Making decisions under uncertainty is a classic topic in operations research and machine learning~\citep{shalev-shwartz_stochastic_2009}. Stochastic optimization formulations explicitly consider uncertainty by optimizing the expected objective over a distribution of unknown parameters. A common approach is the Sample Average Approximation (SAA)~\citep{kleywegt_sample_2002, arjevani_second-order_2020, wang_sample_2024}, which draws many samples from the estimated distribution and solves an approximated deterministic problem minimizing the average cost. While SAA can handle arbitrary uncertainty distributions in theory, it becomes very computationally expensive and still does not consider the distribution during optimization~\citep{kim_guide_2015}. It will lead to optimizing the \emph{sample mean}, which may yield a decision that performs poorly if reality often falls into one of several distinct models far from the mean~\citep{kim_guide_2015, elmachtoub_smart_2022}.

% Our approach differs from theirs by learning a parametric uncertainty model and integrating it into the decision-making loop. Specifically, we train a generative predictor that captures complex uncertainty and directly optimizes the expected decision loss via end-to-end learning. This yields a solution that is tuned to the actual learned distribution of outcomes, without the need for worst-case assumptions.

% On the other hand, robust optimization can protect against worst-case scenarios, but it often assumes a predefined uncertainty set rather than a learned probabilistic model, which has been proven to lead to overly conservative decisions~\citep{ben-tal_robust_2002, beyer_robust_2007}. Distributionally robust optimization (DRO) offers a middle ground by optimizing against the worst-case within an ambiguity set of distributions~\citep{delage_distributionally_2010, wiesemann_distributionally_2014}, yet it requires one to define the set and still returns a single robust solution rather than leveraging a learned probabilistic model of uncertainty. 

\end{document}